\documentclass[]{mosi}

\usepackage{helvet}

\usepackage{amsmath} 
\usepackage[numbers, sort&compress]{natbib}
\usepackage{graphicx}
\usepackage{subcaption} 

\usepackage[toc,page,header]{appendix}
\usepackage[utf8]{inputenc} % allow utf-8 input
\usepackage[T1]{fontenc}    % use 8-bit T1 fonts
\usepackage{hyperref}       % hyperlinks
\usepackage{url}            % simple URL typesetting
\usepackage{booktabs}       % professional-quality tables
\usepackage{nicefrac}       % compact symbols for 1/2, etc.
\usepackage{wrapfig}
\usepackage{titletoc}
\usepackage{minitoc}

\usepackage{array}
\usepackage{etoolbox}

\definecolor{lightblue}{RGB}{200, 230, 255}  
\definecolor{headerblue}{RGB}{150, 200, 255} 

\usepackage{pgfplots}
\pgfplotsset{compat=1.18}
\usepackage{xcolor}
\usepackage{float}
\usepackage{comment}
\usepackage{multirow}
\usepackage{makecell}
\usepackage{siunitx}
\usepackage{tikz}
\usepackage{pgf-pie}
\usepackage[export]{adjustbox}

\usepackage{ragged2e}
\usepackage{tabularx}
\usepackage{caption}
\usepackage{enumitem}
\usepackage{pifont}
\usepackage[hang,flushmargin]{footmisc}

\usepackage{tcolorbox}
\tcbuselibrary{breakable}
\tcbuselibrary{skins}

\usepackage{tabularx}
\usepackage{listings}

\usepackage{threeparttable}
\usepackage{setspace}

\definecolor{MossCyan}{HTML}{82D9FF} 
\definecolor{MossBlue}{HTML}{82B1FF}

\definecolor{ForestGreen}{RGB}{34, 139, 34}
\definecolor{Red}{RGB}{255, 0, 0}
\definecolor{tickG}{rgb}{0.1, 0.588, 0.1}
\definecolor{crossR}{rgb}{0.588, 0.1, 0.1}

\definecolor{frenchblue}{rgb}{0.0, 0.45, 0.73}
\definecolor{babyblue}{rgb}{0.54, 0.81, 0.94}
\definecolor{classicrose}{rgb}{0.98, 0.8, 0.91}
\definecolor{beige}{rgb}{0.96, 0.96, 0.86}
\definecolor{forestgreen}{HTML}{2e7d43}

\definecolor{blue1}{HTML}{91BBE6}
\definecolor{blue2}{HTML}{3F90E0}
\definecolor{blue3}{HTML}{316FAD}

\definecolor{color1}{HTML}{FF9999}
\definecolor{color2}{HTML}{FF6666}
\definecolor{color3}{HTML}{FF3333}
\definecolor{color4}{HTML}{E60000}
\definecolor{color5}{HTML}{B30000}
\definecolor{color6}{HTML}{8CD98C}
\definecolor{color7}{HTML}{53c653}
\definecolor{color8}{HTML}{00B050}
\definecolor{color9}{HTML}{2d862d}
\definecolor{color10}{HTML}{206020}
\definecolor{color11}{HTML}{cca300}

\usepackage{algorithm}
\usepackage{algpseudocode}
\usepackage{cleveref}

\newtcolorbox{promptbox}[2][]{
    colback=white,
    coltext=black,
    arc=3mm,
    boxrule=0.5pt,
    colframe=black!60!white,
    title={#2},
    colbacktitle=black,
    coltitle=white,
    fonttitle=\bfseries,
    top=8pt,
    bottom=8pt,
    left=10pt,
    right=10pt,
    breakable,
    before upper={%
        \linespread{1}\selectfont
        \setlength{\parskip}{1ex plus 0.2ex minus 0.2ex}%
        \setlength{\parindent}{0pt}%
    },
    #1
}

\newtcolorbox{promptblock}{
  colback=blue!2!white,
  colframe=blue!30!gray,
  boxrule=0.6pt,
  arc=3pt,
  left=12pt,
  right=12pt,
  top=8pt,
  bottom=8pt,
  boxsep=0pt,
  before skip=10pt,
  after skip=10pt,
  breakable,
  fontupper=\normalfont,
  parbox=false,
}

\newcommand{\ctext}[1]{\raise0.2ex\hbox{\textcircled{\scriptsize{#1}}}}

\title{MOVA: Towards Scalable and Synchronized Video–Audio Generation}

\author{SII-OpenMOSS Team\textsuperscript{*}}

\abstract{
    Audio is indispensable for real-world video, yet generation models have largely overlooked audio components. Current approaches to producing audio-visual content often rely on cascaded pipelines, which increase cost, accumulate errors, and degrade overall quality. While systems such as Veo 3 and Sora 2 emphasize the value of simultaneous generation, joint multimodal modeling introduces unique challenges in architecture, data, and training. Moreover, the closed-source nature of existing systems limits progress in the field. In this work, we introduce \textbf{MOVA} (\textbf{MO}SS \textbf{V}ideo and \textbf{A}udio), an open-source model capable of generating high-quality, synchronized audio-visual content, including realistic lip-synced speech, environment-aware sound effects, and content-aligned music. MOVA employs a Mixture-of-Experts (MoE) architecture, with a total of 32B parameters, of which 18B are active during inference. 
    It supports IT2VA (Image-Text to Video-Audio) generation task. 
    By releasing the model weights and code, we aim to advance research and foster a vibrant community of creators. The released codebase features comprehensive support for efficient inference, LoRA fine-tuning, and prompt enhancement.
}

\checkdata[Blog]{\url{https://mosi.cn/models/mova}}
\checkdata[Model]{\url{https://huggingface.co/collections/OpenMOSS-Team/mova}}
\checkdata[Code]{\url{https://github.com/OpenMOSS/MOVA}}

\begin{document}
\maketitle
\begingroup
  \renewcommand{\thefootnote}{\fnsymbol{footnote}}
  \setcounter{footnote}{1}
  \footnotetext{Full contributors can be found in the Contributors section.}
\endgroup

\newcounter{num}
\newcommand{\rnum}[1]{\setcounter{num}{#1} \roman{num}}
\crefname{equation}{Eq.}{Eqs.}

% ===== Introduction =====
\section{Introduction}

Video generation has long been a pivotal domain in multimodal generative modeling. Historically, limitations in model capacity, data volume, and scalability hindered these models from achieving practically viable performance. The emergence of the scalable Diffusion Transformer architecture~\citep{DiT} has changed this landscape, giving rise to a series of models-including Sora~\citep{sora}, OpenSora~\citep{OpenSora,OpenSora2}, Wan~\citep{wan}, and LTX~\citep{LTX-Video}—that can generate high-fidelity, realistic videos. Beyond basic video synthesis, these models also demonstrate advanced capabilities such as long video generation~\cite{yang2025longlive}, controllable synthesis~\cite{cai2024ditctrl}, and even few-shot learning and visual reasoning~\citep{Veo3_Reasoner_Learner,tong2025thinking}.

% pxy
However, traditional video generation models often neglect the audio component, despite its critical role in multimedia content. Producing videos with synchronized sound typically requires a cascaded pipeline\cite{gao2025wans2vaudiodrivencinematicvideo}, e.g., generating video first and then synthesizing audio based on the visuals, or vice versa. Such pipelines inherently limit generation quality, as the audio and video modalities do not interact during synthesis. End-to-end models like Veo3~\citep{google-veo3} have demonstrated that high-fidelity audio-visual generation with precise synchronization is possible, highlighting the importance of joint audio-video modeling. Similarly, Sora2~\citep{sora22025} exhibits impressive capabilities in generating synchronized audio and video. Yet, all these state-of-the-art systems are closed-source, and subsequent releases such as Wan2.6~\citep{wan2.6}, Kling Video 3,0~\citep{kling3.0}, and Seedance 2.0~\citep{seedance2.0} remain proprietary. As a result, the development of high-quality video-audio generation models remains underexplored in the research community.

% pxy
Compared to video-only generation models, video-audio generation models introduce three major challenges:
\textbf{1. Data Pipeline:} Incorporating the audio modality requires a fine-grained audio-video captioning pipeline.
\textbf{2. Modality Fusion:} To achieve harmonious and synchronized video-audio generation, these two modalities must mutually integrate information during the generation process.
However, simultaneous bi-modal generation faces challenges due to disparity in native information density between the two modalities, as well as challenges regarding fusion mechanisms and efficiency.
\textbf{3. Scalability Verification:} Most existing open-source models are evaluated on small-scale architectures and limited datasets~\citep{wang2025universe,low2025ovi,hu2025harmony,zhang2025uniavgen}, often falling short in achieving high-quality results across both modalities. It remains an open question whether video-audio models can sustain continuous performance improvements with larger datasets and model scales.

\begin{figure}[t]
    \centering
    \includegraphics[width=\textwidth]{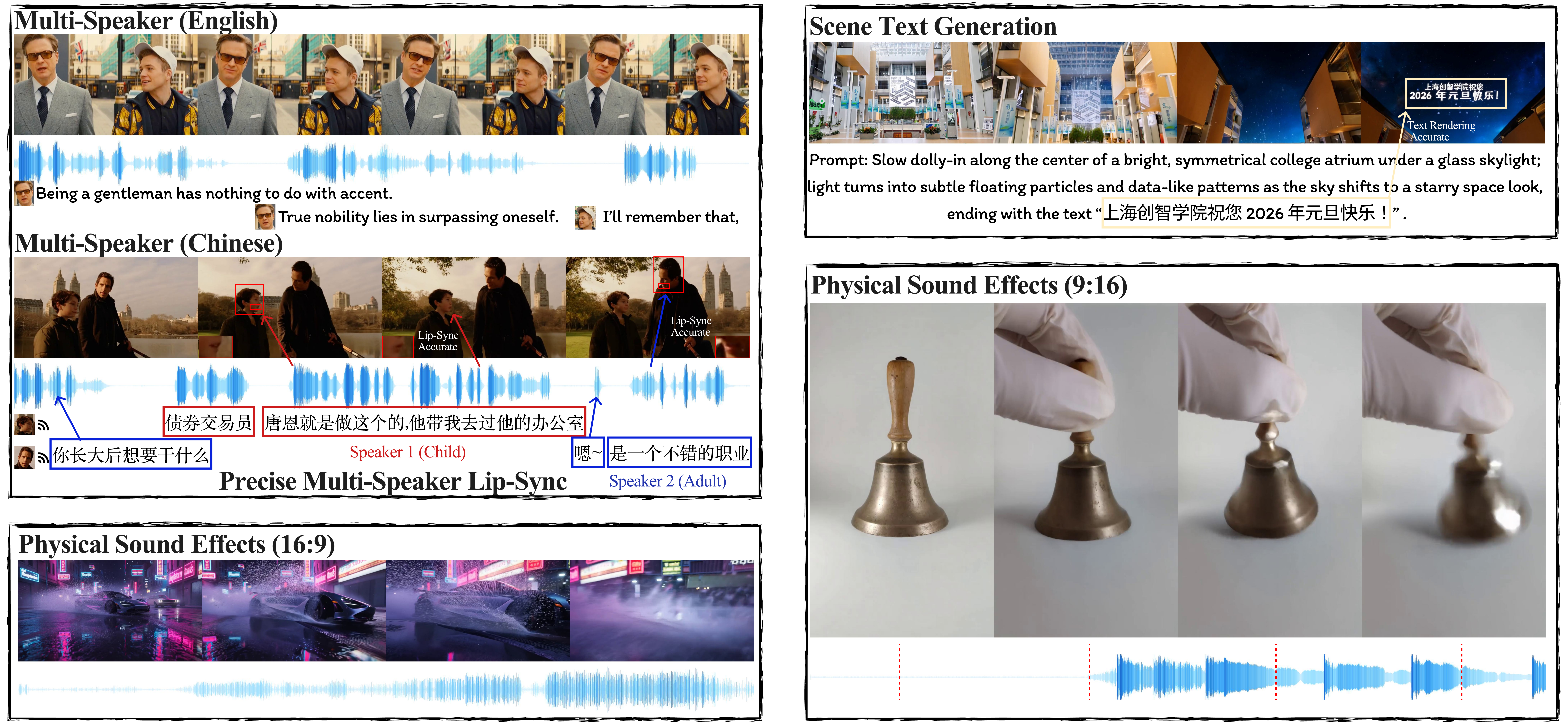}
    \caption{Overview of MOVA capabilities. MOVA generates synchronized video and audio across diverse scenarios: multi-speaker speech with precise lip synchronization in both English and Chinese, physical sound effects aligned with visual events, and scene text generation. The model supports both 16:9 and 9:16 aspect ratios.}
    \label{fig:teaser}
\end{figure}

% pxy
In this work, we present \textbf{MOVA} (\textbf{MO}SS \textbf{V}ideo and \textbf{A}udio), a video-audio generation foundation model capable of synthesizing multilingual speech with high-quality lip synchronization, as well as environmental sounds with precise audio-visual alignment. To achieve high-quality video-audio generation, we propose an asymmetric dual-tower architecture, combining a pre-trained video tower with a pre-trained audio tower. For modality fusion, we employ a bidirectional cross-attention mechanism that enables rich interaction between the two modalities. 
To unleash the potential of this architecture, we construct a fine-grained video-audio annotation pipeline and scale up the training data, resulting in strong performance in synchronized video-audio generation. We observe consistent improvements in both lip synchronization and video-audio alignment metrics as training progresses. As illustrated in Figure~\ref{fig:teaser}, MOVA realizes scalable and synchronized video-audio generation. To support research and foster community creation, we release all model weights along with training, inference, and fine-tuning code.

Our contributions can be summarized as follows:
\begin{itemize}
    \item We develop \textbf{MOVA}, a synchronized video-audio generation model with an asymmetric dual-tower architecture that leverages pre-trained video and audio generation models.
    \item We design a fine-grained audio-video captioning pipeline to produce high-quality bimodal training data at scale.
    \item By scaling video-audio training, we achieve continuous improvements in synchronization performance across both modalities.
\end{itemize}

% ===== Model Architecture =====
\section{Model Architecture}
\label{sec:architecture}

\subsection{Preliminaries}
\label{sec:prelim}
We consider joint generation of temporally aligned video-audio pairs from a text prompt (optionally with a first-frame image for I2VA). Let $V \in \mathbb{R}^{(1+T^v)\times H \times W \times 3}$ be a video of $T^v+1$ frames and $A \in \mathbb{R}^{T^a \times 1}$ be the corresponding 48\,kHz mono audio waveform. We adopt pretrained VAEs to define compact latent spaces: the Wan2.1 video VAE~\citep{wan} compresses $V$ into a spatiotemporal latent $x^v$, and a DAC-style audio VAE from HunyuanVideo-Foley~\citep{shan2025hunyuanvideofoleymultimodaldiffusionrepresentation} encodes $A$ into an audio latent $x^a$. All subsequent modules operate in these latent spaces. Given a condition $c$, our goal is to generate synchronized $(x^v, x^a)$.

\begin{figure}
  \centering
  \includegraphics[width=\linewidth]{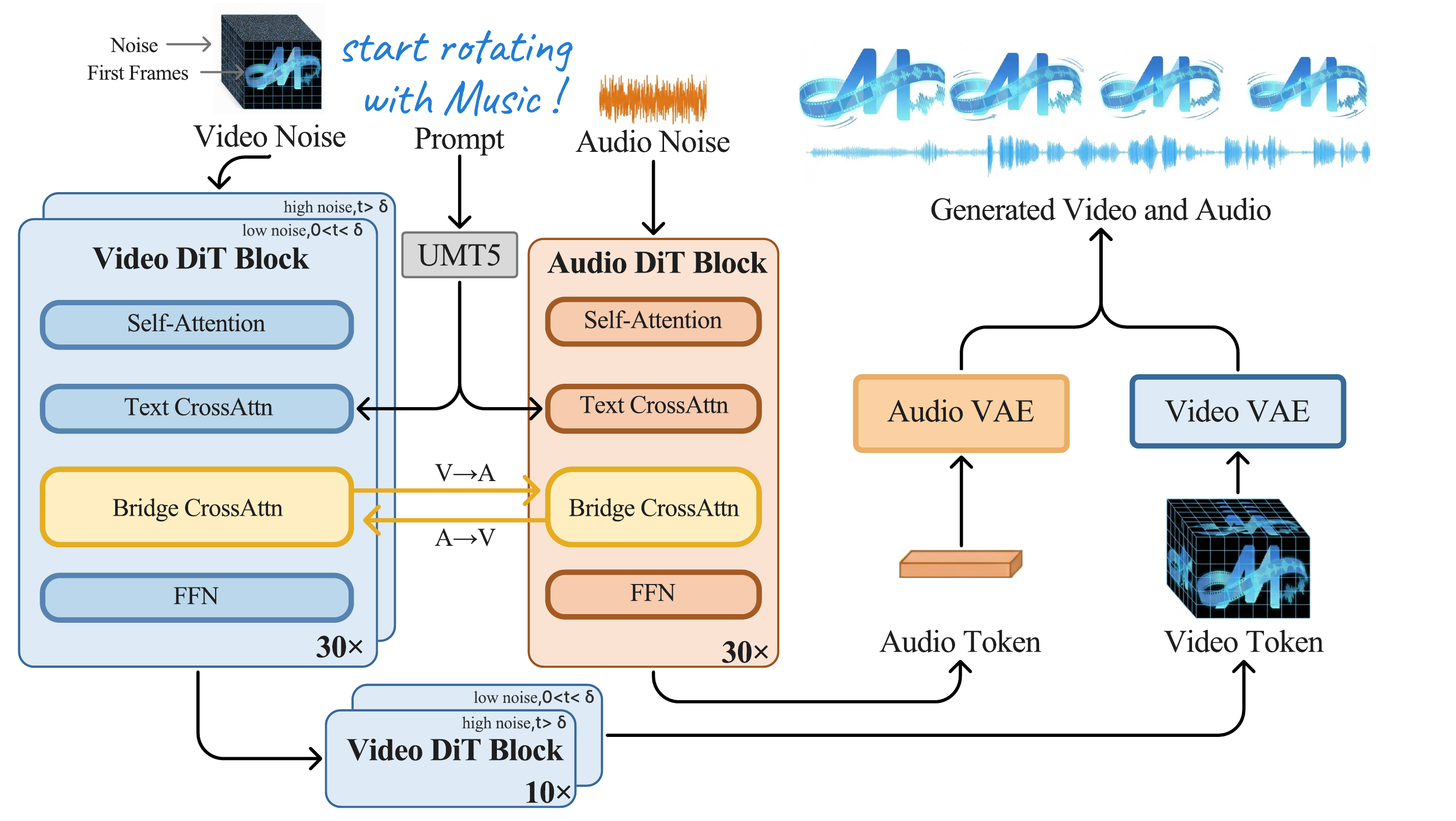}
  \label{fig:overview}
  \caption{Model Structure Overview. MOVA couples an A14B video DiT backbone and a 1.3B audio DiT backbone via a 2.6B bidirectional Bridge module. }
\end{figure}

Our training objective follows the flow matching method~\citep{lipman2022flow,liu2022flow}. At each timestep $t$,
\begin{align}
x_t^v &= (1-t) x^v + t \varepsilon^v, \nonumber \\
x_t^a &= (1-t) x^a + t \varepsilon^a,
\end{align}
where $\varepsilon^v, \varepsilon^a \sim \mathcal{N}(0, I)$ and $t \sim \mathcal{U}(0,1)$.

The target velocity is given by the derivative of the interpolation path:
\begin{align}
v_t^v &:= \frac{\mathrm{d}x_t^v}{\mathrm{d}t} = \varepsilon^v - x^v, \nonumber \\
v_t^a &:= \frac{\mathrm{d}x_t^a}{\mathrm{d}t} = \varepsilon^a - x^a.
\end{align}

The training loss is defined as a standard flow matching objective:
\begin{equation}
\mathcal{L}_{\mathrm{FM}} = \mathbb{E}_{t, c, x^v, x^a, \varepsilon}
\Big[
\lambda_v \left\| \hat{v}^v_\theta(x_t^v, x_t^a, t, c) - v_t^v \right\|_2^2
\;+\;
\lambda_a \left\| \hat{v}^a_\theta(x_t^v, x_t^a, t, c) - v_t^a \right\|_2^2
\Big],
\end{equation}
where $\theta$ denotes the model parameters, $\hat{v}_\theta$ denotes the predicted cross-modal velocity field, $c$ denotes the conditioning signal (text, optionally augmented with an input image), and $\lambda_v,\lambda_a$ are loss weights that balance the video- and audio-velocity regression terms, respectively.

\subsection{Dual-Tower Diffusion Transformer}

Our goal is to leverage powerful pretrained single-modality diffusion models and achieve synchronized video--audio generation with minimal additional cost. Specifically, we adopt Wan2.2 I2V A14B~\citep{wan} as the video backbone (for image+text conditioned I2VA) and a 1.3B text-to-audio diffusion model with a Wan2.1-style architecture as the audio backbone. We couple these two backbones through a lightweight dual-tower conditional Bridge, enabling bidirectional information exchange while keeping the core towers intact.

\paragraph{Bridge Module.}
The bridge operates at the hidden-state level of the two DiT backbones. At each interaction layer, two cross-attention blocks are added: one injects video hidden states into the audio DiT. and one injects audio hidden states into the video DiT.

\paragraph{Aligned RoPE.}
Video and audio latents live on different temporal grids: video latents are relatively coarse (due to temporal downsampling in the video VAE), while audio latents are much denser. If we apply cross-attention with naive positional indices, tokens that represent the same physical time can be assigned to different positional slots (and conversely different physical times can share nearby indices), so queries and keys may correspond to mismatched physical times, causing audio--visual drift; this positional mismatch also breaks the translation invariance of the interaction process.

Similarly to MMAudio~\citep{cheng2025mmaudio} and OVI~\citep{low2025ovi}, we modify standard RoPE~\citep{su2021roformer} to align the two modalities on the same time grid.:
\[
\tilde{\mathbf{q}}_{(m)} = R\!\left(\tfrac{p}{\theta_m}\right)\mathbf{q}_{(m)},\quad
\tilde{\mathbf{k}}_{(m)} = R\!\left(\tfrac{p}{\theta_m}\right)\mathbf{k}_{(m)},\qquad
R(\phi)=\begin{bmatrix}\cos\phi&-\sin\phi\\ \sin\phi&\cos\phi\end{bmatrix}.
\]

Let $f_v$ and $f_a$ denote the latent frame rates of video and audio after their respective VAEs. We map video indices into the audio time units by scaling the temporal position with the ratio $s = f_a / f_v$:
\[
p_v(i) = s \cdot i, \qquad p_a(j) = j.
\]
This puts video and audio tokens on the same temporal scale.

% ===== Data Engineering =====
\section{Data Engineering}
\label{sec:data}

\begin{figure}[!t]
  \centering
  \includegraphics[width=\linewidth]{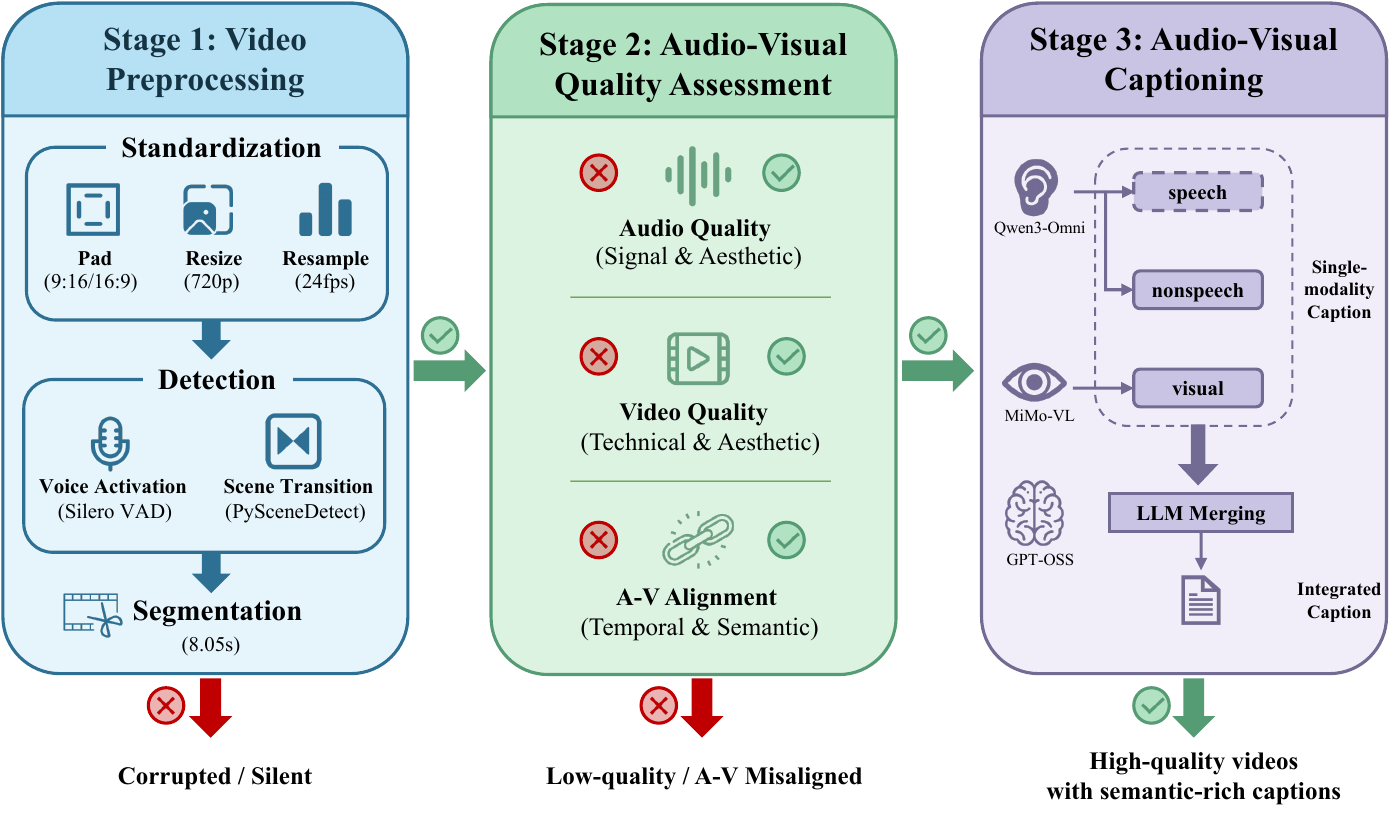}
  \caption{Data curation overview. Our data pipeline consists of three stages. In the first stage, we preprocess the raw data into fixed-length clips with a resolution of 720p, a frame rate of 24fps, and a duration of 8.05s. In the second stage, we filter these clips based on audio quality, video quality, and audio-visual alignment to obtain high-quality, synchronized clips. In the third stage, we utilize Qwen3-Omni and MiMo-VL to label the audio and visual information within the videos, respectively, and finally use GPT-OSS to merge these single-modality captions. Through our data pipeline, we have successfully curated high-quality audio-visual content along with corresponding, semantically rich captions.}
  \label{fig:data_pipeline}
\end{figure}

\begin{table}[!b]
\centering
\setlength{\tabcolsep}{8pt}
\renewcommand{\arraystretch}{1.2}
\caption{Retention Ratio of Total Dataset Duration}
\resizebox{\columnwidth}{!}{%
\begin{tabular}{lcccc}
\toprule
  & Raw & Stage 1 (speech+nonspeech) & Stage 1 (speech only) & Stage 2  \\ 
  \midrule
\begin{tabular}{@{}l@{}}Retention Ratio \\ (Relative to Raw Video) \end{tabular} 
& 100\% & 84.57\% & 58.75\% & 26.39\% \\ 
\bottomrule
\end{tabular}
}
\label{tab:drr}
\end{table}

\subsection{Overview}
To transform heterogeneous and noisy raw videos into reliable resources for video-audio generation, we develop a systematic data curation pipeline. The pipeline first cleans and standardizes the inputs, followed by the annotation of high-quality, richly-detailed captions across multiple modalities. By carefully structuring the data pipeline into three stages, we progressively filter out low-quality samples to retain high-fidelity data characterized by strong audio-visual consistency and coherent semantic labels.

We curate a high-quality subset of audio-visual data from various public datasets (data collection, \Cref{sec:data_collection}). Our collection spans multiple video formats (e.g., movies, vlogs, and animations) and diverse themes ranging from education and sports to animation and interviews. The three-stage pipeline then processes the raw video data. In the first stage (video preprocessing, \Cref{sec:preprocessing}), we normalize aspect ratios, resize videos, resample streams, and segment videos into fixed-length clips based on VAD and scene transition annotations. In the second stage (audio-visual quality assessment, \Cref{sec:quality_assess}), we evaluate audio quality, video quality, and audio-visual alignment, retaining only clips that are both high-fidelity and well-aligned. In the third stage (audio-visual captioning, \Cref{sec:caption}), we design structured and instructive prompts, generate modality-specific captions with MLLMs, and integrate them into unified annotations using a LLM. Our data pipeline is presented in \Cref{fig:data_pipeline} and retention ratio is shown in \Cref{tab:drr}.

\subsection{Data Collection}
\label{sec:data_collection}
We use filtered HQ subsets of the following public video datasets in our work:  VGGSound~\citep{chen2020vggsound}, AutoReCap~\citep{haji2024taming}, ChronoMagic-Pro~\citep{yuan2024chronomagic}, ACAV-100M~\citep{ACAV100M}, OpenHumanVid~\citep{li2025openhumanvid}, SpeakerVid-5M~\citep{zhang2025speakervid}, and OpenVid-1M~\citep{nan2024openvid}.
In addition, we utilize a large amount of in-house data.
Our dataset encompasses a broad spectrum of domains (education, sports, and beauty, news, etc.), providing the distributional diversity necessary to enhance the model's generalization across complex real-world scenarios.

\subsection{Video Preprocessing}
\label{sec:preprocessing}
We design and implement a scalable video preprocessing pipeline based on the Ray distributed framework, balancing data quality and processing efficiency. Initially, raw video data is filtered to remove samples with decoding failures or missing valid audio channels. Videos with unconventional container or encoding formats are remuxed or transcoded, respectively. Then, the pipeline generates segmentation metadata through three sequential steps:
\begin{itemize}
    \item \textbf{Core Content Normalization}: The FFmpeg cropdetect filter is applied to detect blank areas and retain the core visual content. The main content is then centered, resized to a 720p resolution, and symmetrically padded with black borders (pillarboxing or letterboxing) as necessary to conform to a 9:16 or 16:9 aspect ratio.
    \item \textbf{Voice Activity Detection}: The audio track is extracted from each video and analyzed by the Silero Voice Activity Detection (VAD) \cite{SileroVAD} model to identify speech and non-speech intervals.
    \item \textbf{Scene Transition Analysis}: PySceneDetect is employed to detect and record the timestamps of scene change points throughout the video.
\end{itemize}
By integrating the temporal information from VAD and scene transition analysis, we generate four types of fixed-duration (8.05-second) video segments: single-scene speech, single-scene non-speech, multi-scene speech, and multi-scene non-speech. This duration corresponds precisely to 193 video frames at 24 fps, calculated as the initial frame plus 8 seconds of video (1 + 8 × 24). For speech segments, the start time is adaptively adjusted to avoid truncating ongoing speech and ensure spoken-content continuity; the detailed algorithm is provided in \Cref{app:window_algorithm}. Ultimately, only speech segments are selected for training, accounting for 69.47\% of all preprocessed segments.

\subsection{Audio-Visual Quality Assessment}
\label{sec:quality_assess}
We conduct data quality assessment along three main dimensions: audio quality, video quality, and audio-visual alignment.

\begin{itemize}
    \item \textbf{Audio quality}: We compute signal-level metrics such as silence ratio and bandwidth, and further evaluate both signal and aesthetic aspects using the Audiobox-aesthetics audio quality assessment tool \cite{tjandra2025meta}.
    
    \item \textbf{Video quality}: We apply the DOVER video quality assessment tool \cite{wu2023dover} to assess the video from both technical and aesthetic perspectives.
    
    \item \textbf{Audio-visual alignment}: We employ SynchFormer \cite{iashin2024synchformer} to compute the temporal audio-visual synchronization of each video, and use ImageBind \cite{girdhar2023imagebind} to evaluate semantic audio-visual alignment.
\end{itemize}

To determine the filtering thresholds, we manually inspect the videos retained under different metric cutoffs and set reasonable thresholds for each dimension accordingly. In addition, we apply an audio classification model \cite{chen2024eat} to categorize audio and construct speech/non-speech subsets depending on the target capability (e.g., lip synchronization vs.\ general foley/ambience modeling). We provide the detailed filtering thresholds in \Cref{app:thresholds}.

\subsection{Audio-Visual Captioning}
\label{sec:caption}
We employ open-source models for audio-visual captioning and subsequently use a large language model (LLM) to merge the generated captions into coherent natural language descriptions, utilizing both NVIDIA GPUs and Ascend NPUs.

\paragraph*{Bimodality Model.}
To annotate the filtered audio-visual contents, we employ distinct pipelines for video and audio modalities. For video annotation, we utilize the MiMo-VL-7B-RL model \cite{Yue2025MiMoVLTR} to generate video descriptions, with explicit instructions focusing on video scene transitions. For audio annotation, we implement a dual-model strategy to separately handle speech and non-speech components. Specifically, Qwen3-Omni-Instruct \cite{Qwen3-Omni} was used for speech transcription, while Qwen3-Omni-Captioner \cite{Qwen3-Omni} was applied to generate captions for non-speech sound and music. We then integrate these two subsets of annotations. This joint annotation strategy enables comprehensive coverage of both linguistic content and acoustic characteristics, reducing information loss and capturing multi-aspect audio semantics.
\paragraph*{Caption Merging.}
For annotations generated by the separate modality pipelines, our primary goal is to integrate the content from the visual and audio dimensions. We employ the GPT-OSS-120B model \cite{agarwal2025gpt} to merge the video captions with the aggregated audio annotations (comprising both speech and non-speech elements) while performing a consistency check to resolve potential cross-modal conflicts. Specifically, the model verifies the alignment between visual scenes and audio events to resolve potential conflicts. It then synthesizes these inputs into a cohesive, natural language description that seamlessly blends visual information with audio details, ensuring the final output is contextually unified and suitable for the target application. All prompts of our caption process and a detailed caption example are provided in \Cref{app:caption}.

% ===== Training Strategy =====
\section{Training Strategy}
\label{sec:training}

\subsection{Overview}
\begin{figure}[!t]
  \centering
  \includegraphics[width=\linewidth]{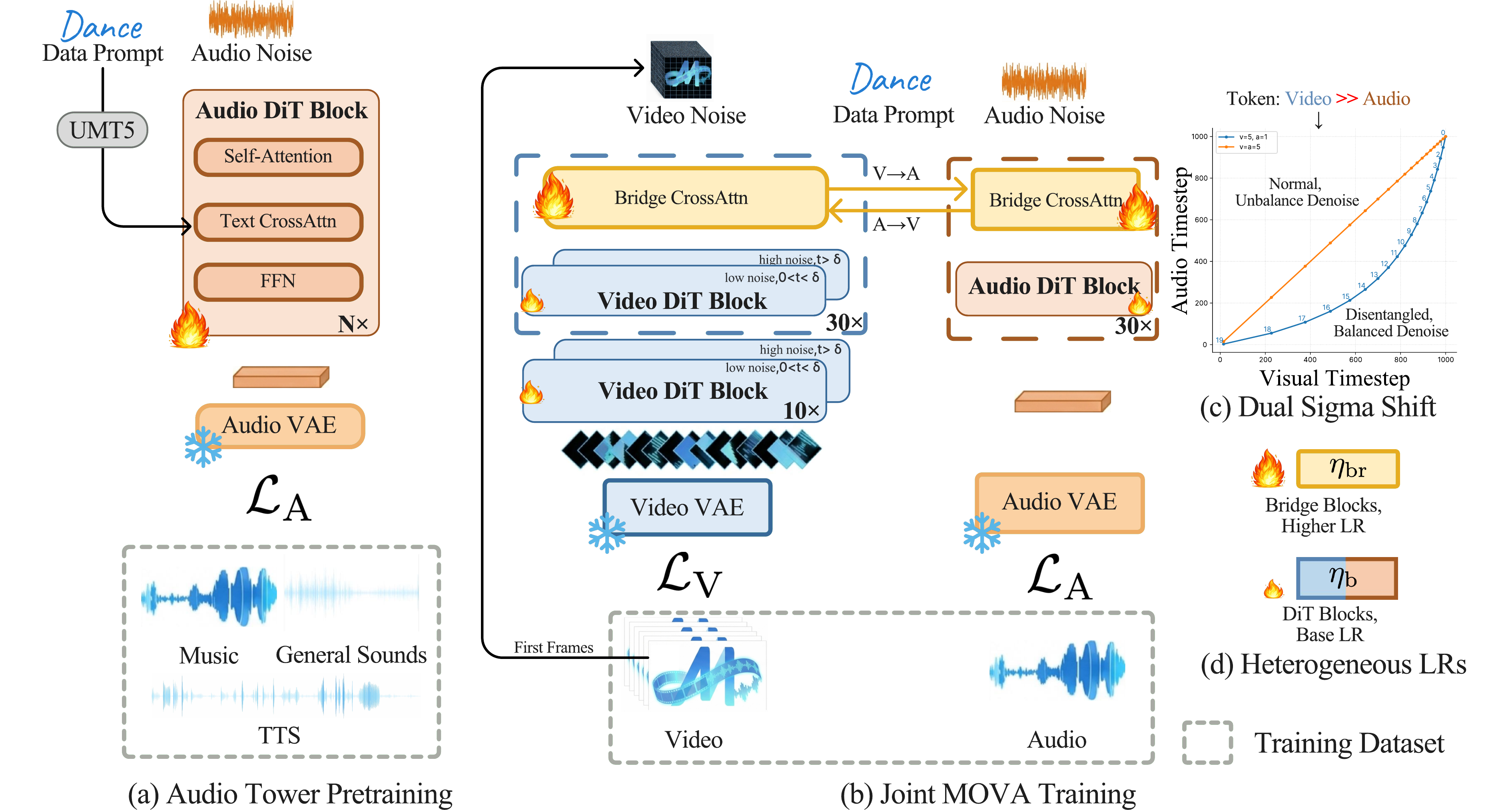}
  \caption{\textbf{Training pipeline overview.} 
  (a) Audio tower pretraining: We train a 1.3B text-to-audio model with Wan2.1-style architecture on music, general sounds, and TTS data. The audio VAE remains frozen during this stage.
  (b) Synchronous joint training: The video tower (A14B, blue) and audio tower (1.3B, orange) are connected via bidirectional Bridge cross-attention modules. (c) Video and audio timesteps are sampled independently, allowing each modality to follow its own noise schedule. (d) Bridge modules use a higher learning rate ($\eta_{\text{br}}=2\times10^{-5}$) than backbone DiT blocks ($\eta_{\text{b}}=1\times10^{-5}$) to accelerate cross-modal alignment while preserving pretrained priors. Both VAEs remain frozen throughout training.}
  \label{fig:training_pipeline}
\end{figure}

Our training pipeline consists of two stages, as illustrated in Figure~\ref{fig:training_pipeline}. We initialize the video tower and video VAE from Wan2.2 weights~\citep{wan}, and adopt the audio VAE from HunyuanVideo-Foley~\citep{shan2025hunyuanvideofoleymultimodaldiffusionrepresentation}; both VAEs remain frozen throughout training. 
In the first stage (audio tower pretraining, \Cref{sec:audio_tower_training}), we train a standalone 1.3B text-to-audio DiT on diverse audio data covering music, general sounds, and speech. In the second stage (joint training, \Cref{sec:hetero_lr}), the pretrained video tower (A14B), audio tower, and a randomly initialized Bridge module are optimized together with heterogeneous learning rates. This joint training further proceeds through three phases with progressively refined data and resolution (\Cref{sec:staged_training}): Phase 1 (360p, diverse data), Phase 2 (360p, quality-filtered data), and Phase 3 (720p, highest-quality data).

\subsection{Audio Tower Training} 
\label{sec:audio_tower_training}

\paragraph{Overview.}
To aligned with the video tower, we train an audio tower with the same architecture of Wan2.1-1.3B backbone. Relative to the original Wan2.1 setup, we replace Wan2.1's 3D positional encoding over $(f,h,w)$ with a 1D positional encoding along the temporal axis.
 All other components and depth remain unchanged to maximally reuse engineering and training practices.

\paragraph{Training Data.}
The text-to-audio model is trained on three domains: (i) general sounds drawn from WavCaps and VGGSound~\citep{mei2023wavcaps,chen2020vggsound}; (ii) music from JamendoMaxCaps~\citep{roy2025jamendomaxcaps}; and (iii) our in-house TTS data. We train on fixed-length clips. Each clip is paired with a text prompt, which also carries an explicit duration token to control target length.

\paragraph{Evaluation Metrics.}We assess the capability of our audio tower through various robustness metrics and compare with other size-matched baselines. Specifically, we consider fidelity, diversity, semantic alignment, and perceptual quality, which provide an overall view of audio generation performance. 
For fidelity, we use the Fréchet Distance (FD) with OpenL3 Embeddings~\citep{openl3}, which measures how closely the generated audio matches real audio in distribution. Diversity is quantified using the Inception Score (IS)\citep{is} which evaluates whether the model produces both varied and high-quality samples. Furthermore, robustness is captured by KL divergence with PaSST\citep{passt} in order to assess the stability of generated audio across different conditions. To assess the semantic consistency, we compute the CLAP score\citep{clap} to measure alignment between the conditioning text prompt and the generated audio.
Beyond four metrics mentioned above, we incorporate perceptual evaluation using the AudioBox Aesthetic Benchmark\citep{audiobox}. The benchmark reports four human—preference-oriented indicators: aesthetics, naturalness, intelligibility and consistency. These dimensions capture overall appeal, absence of audible artifacts, clarity of linguistic or musical content, and the temporal coherence across the entire audio clip respectively.

\paragraph{Results.}
As shown in Table~\ref{tab:tta}, our model achieves competitive performance on the AudioCaps~\citep{kim-NAACL-HLT-2019} benchmark. It attains a strong IS of 10.54, clearly higher than AudioLDM2 (7.79), while maintaining a competitive CLAP score of 0.463. In terms of FD openl3, our result (72.25) is nearly identical to AudioLDM2~\citep{liu2023audioldm,audioldm2-2024taslp} (72.04) and much better than TangoFLUX~\citep{hung2024tangoflux} (80.47), indicating stronger semantic fidelity. The KL divergence (1.47) also improves upon AudioLDM2 (1.66), suggesting better distribution alignment.

In Table~\ref{tab:audiobox}, our method achieves the best results on Consistency (CU = 5.56) and Perceptual Quality (PQ = 6.20), surpassing advanced baselines such as AudioLDM2 and Tango2~\citep{majumder2024tango2}. Although it does not lead in CE or PC, the scores remain competitive. All these results demonstrate that our audio tower provides improvements in fidelity and coherence, while enhancing perceptual quality and semantic alignment compared to existing approaches.

\begin{table}[t]
\centering
\setlength{\tabcolsep}{8pt}
\renewcommand{\arraystretch}{1.2}
\caption{Text-to-audio performance benchmark with AudioCaps Dataset. We report parameter size, number of function evaluations (NFE), Fréchet Distance with Openl3 embeddings (FD openl3), KL divergence with PaSST (KL passt), CLAP score, and Inception Score (IS).}
\begin{tabular}{lcccccc}
\toprule
Model & Params & NFE & FD openl3$\downarrow$ & KL passt$\downarrow$ & CLAP score$\uparrow$ & IS$\uparrow$ \\ \midrule
TangoFLUX~\citep{hung2024tangoflux}       & 516M  &  50 & 80.47 & 1.02  & 0.546 & 13.28 \\
AudioLDM2~\citep{audioldm2-2024taslp}       & 346M  & 200 & 72.04 & 1.66  & 0.409 &  7.79 \\
Ours & 1.3B  & 100 & 72.25 & 1.47  & 0.463 & 10.54 \\ \bottomrule
\end{tabular}
\label{tab:tta}
\end{table}

\begin{table}[t]
\centering
\setlength{\tabcolsep}{8pt}
\renewcommand{\arraystretch}{1.2}
\caption{AudioBox Benchmark results. CE, CU, PC, and PQ denote the four evaluation indicators from AudioBox. Best results are highlighted in bold.}
\begin{tabular}{lcccc}
\toprule
Model & CE & CU & PC & PQ \\ \midrule
AudioLDM~\citep{liu2023audioldm}        & 3.27 & 5.10 & 3.23 & 5.82 \\
AudioLDM2~\citep{majumder2024tango2}       & 3.48 & 5.54 & 3.00 & 6.09 \\
Make-An-Audio 2~\citep{huang2023makeanaudio} & 3.23 & 4.98 & 3.17 & 5.58 \\
Tango 2~\citep{majumder2024tango2}         & 3.47 & 5.20 & \textbf{3.84} & 5.89 \\
TangoFLUX~\citep{hung2024tangoflux}       & \textbf{3.54} & 5.07 & 3.64 & 5.78 \\
Ours            & 3.41 & \textbf{5.56} & 3.04 & \textbf{6.20} \\ \bottomrule
\end{tabular}
\label{tab:audiobox}
\end{table}

\subsection{Progressive Joint Training} 
\label{sec:staged_training}

Our joint training follows a three-phase data and resolution curriculum, progressively refining both data quality and output resolution.

\textbf{Phase 1 (360p baseline):} We initialize from pretrained Wan2.2 A14B (video) and a 1.3B audio tower, inserting the Bridge module with random initialization. Training proceeds at 360$\times$640 resolution for 193 frames (8 seconds at 24 fps) on 1024 GPUs. We train on approximately 61,500 hours of diverse video-audio data drawn from SpeakerVid5M, Chinese drama, cartoon, movies, YouTube, and OpenHumanVid. We use asymmetric sigma-shift values with $\text{shift}_v=5.0$ and $\text{shift}_a=1.0$ to focus video learning on aggressive denoising while keeping audio transitions smoother, and employ aggressive text dropout ($p_{\text{drop}}^{\text{text}}=0.5$) to force Bridge-based alignment learning. This phase runs for 1 epoch (15 days).

\textbf{Phase 2 (quality-filtered alignment):} We refine the noise schedule by aligning audio to match video, setting $\text{shift}_a=5.0$ to match the video schedule. 
While Phase~1 uses a smoother audio schedule for stability, we find that high-fidelity timbre is sensitive to the noise schedule and denoising steps. Therefore, in Phase~2 we align the audio sigma shift to the video setting ($\text{shift}_a=5.0$) to strengthen audio denoising and improve timbre fidelity.
This alignment enables the audio tower to benefit from the same aggressive noise schedule that improves video quality, without requiring architectural changes---only the sigma-shift parameter is updated. We also reduce text dropout to $p_{\text{drop}}^{\text{text}}=0.2$ to allow text-guided refinement while retaining learned cross-modal priors, and introduce LUFS normalization to mitigate CFG-induced loudness explosion. This phase trains on approximately 37,600 hours of quality-filtered data for 1 epoch (7 days).
To improve training quality, we curate the Phase 2 dataset using three complementary filters. First, we use OCR to identify videos without burned-in subtitles, retaining $\sim$9.5M clips and appending ``\texttt{This video has no subtitles.}'' to their prompts to teach the model this distinction. Second, we retain videos with LSE-D $\leq 9.5$ and LSE-C $\geq 4.5$, yielding $\sim$2.5M clips with high-quality lip-audio correspondence. Third, we apply DOVER technical quality score $>0.15$ to select videos with superior visual fidelity, yielding $\sim$2.4M clips. The resulting dataset contains 16.8M clips ($\sim$37,600 hours), balancing scale and quality.

\textbf{Phase 3 (720p fine-tuning):} We upscale to 720$\times$1280 resolution, training on approximately 11,000 hours of the 720p highest-quality subset (DOVER technical score $>0.14$). The increased sequence length requires modified parallelism configuration: we increase context parallelism from CP=8 to CP=16. Checkpoint frequency increases to every 2000 steps to capture rapid convergence. This phase runs for 1 epoch (20 days).

\paragraph{Computational Resources.}
All three phases run on 1024 GPUs (128 nodes, 8 GPUs per node). For 360p training (Phases 1--2), we use CP=8, yielding effective batch size 128. For 720p fine-tuning (Phase 3), increased sequence length requires CP=16, reducing effective batch size to 64. The complete training spans 42 days, totaling approximately 43,000 GPU-days. See Appendix~\ref{app:training_hparams} for the complete hyperparameter configuration.

\subsection{Optimization Details}
\label{sec:hetero_lr}

Throughout all three phases, we optimize the full model end-to-end: the Bridge module and both pretrained towers are updated jointly from the first step. This differs from a two-stage warm-start that first trains the Bridge with frozen towers and then fine-tunes the full model. In our early experiments, the two-stage scheme reached an early performance plateau, which motivated end-to-end joint optimization. The main tension is that the Bridge must learn cross-modal correspondence quickly, while the large pretrained towers should remain stable and preserve their strong unimodal priors. We mitigate this tension with heterogeneous learning rates across module groups.

\paragraph{Heterogeneous Learning Rates.}
To balance fast Bridge convergence with tower stability, we use a higher learning rate for the Bridge ($\eta_{br} = 2 \times 10^{-5}$) than for the backbone towers ($\eta_b = 1 \times 10^{-5}$). This factor-of-two difference accelerates Bridge learning while reducing forgetting in the pretrained towers. In our experiments, a uniform learning rate either destabilizes the towers (if too high) or leaves the Bridge under-trained (if too low).

\paragraph{Dual Sigma Shift.}
\label{par:dual-sigma-shift}
Conventional joint diffusion that forces video and 
audio to share the same timestep can be suboptimal. The two modalities have different effective complexities: audio uses fewer tokens per second, yet timbre fidelity is highly sensitive to noise schedules. A single noise level may be too aggressive for one modality and too mild for the other, causing imbalanced gradients.

To enable this, we decouple the timestep sampling for each modality. During training, we independently draw $t_v$ and $t_a$ from $\mathcal{U}(0,1)$:
\begin{equation}
\begin{aligned}
z_v^{t_v} &= (1-\sigma_v(t_v)) \cdot z_v^0 + \sigma_v(t_v) \cdot \epsilon_v, \\
z_a^{t_a} &= (1-\sigma_a(t_a)) \cdot z_a^0 + \sigma_a(t_a) \cdot \epsilon_a,
\end{aligned}
\end{equation}
where $\sigma_m(t) = \frac{\text{shift}_m \cdot t}{\text{shift}_m + t(1-\text{shift}_m)}$ controls the noise schedule for modality $m \in \{v, a\}$.

This decoupling provides two benefits. First, each modality follows its natural denoising trajectory---we set $\text{shift}_v=5.0$ (aggressive) and $\text{shift}_a=1.0$ (gradual) in Phase 1, then align them to $\text{shift}_a=5.0$ in Phase 2 to improve timbre fidelity. Second, noise schedules can be adjusted at inference time without retraining.

\subsection{Training Efficiency}
\label{sec:training_eff}

For large-scale training, we shard model parameters with Fully Sharded Data Parallel (FSDP)~\citep{zhao2023pytorch} and adopt sequence parallelism via USP~\cite{fang2024usp}, achieving approximately 35\% MFU. 
To eliminate redundant VAE computation introduced by sequence parallelism, we follow the approach in Wan~\cite{wan}: for each CP group, the input preprocessing (primarily the VAE) is performed only once per CP step, and the preprocessed features are then broadcast from a designated rank to all other ranks within the same CP group before being fed into the DiT backbone, thereby avoiding duplicated computation.
We use manual memory management to avoid the same Python's garbage collection (GC) overhead reported in OpenSora2~\cite{OpenSora2}.
We also port our training stack to Ascend NPUs and apply operator fusion for attention kernels, tensor layout transforms, and rotary embedding computation to reduce framework overhead. Under an 8-device configuration (CP=4, DP-shard=2), we measure 34.1~s/step on $8\times$ Ascend 910A2; benchmark details are provided in Appendix~\ref{app:ascend_benchmark}.

Because standard FSDP requires a consistent computation graph, for the A14B MoE video tower we adopt an alternating optimization strategy: on odd steps we sample high-noise timesteps for all samples and optimize the high-noise DiT, while on even steps we sample low-noise timesteps and optimize the low-noise DiT. In addition, the shared bridge and the audio tower are optimized at every step.

% ===== Efficiency Adaption =====
% \section{Efficiency Adaption}
% \label{sec:efficiency}

% \input{chapters/Efficiency}

% ===== Generation Workflow =====
\section{Inference}
\label{sec:inference}

% \subsection{Inference Pipeline} %这边只放主线的pipeline  cfg放到discussion

\subsection{Dual Classifier-Free Guidance}
% \label{subsec:infer guide}
% \paragraph{Dual Classifier-Free Guidance.}
In the text-conditioned video-audio generation (T2VA) setting, we may view a paired video-audio sample as a single joint latent, where the only explicit condition is text. However, from a single-modality perspective, the other modality provides additional conditional information: when predicting video, audio serves as a condition, and vice versa. This perspective introduces extra controllability, because we can separately adjust the guidance strengths for text conditioning and cross-modal conditioning.

Following the dual Classifier-Free Guidance (dual CFG) proposed in InstructPix2Pix~\citep{brooks2023instructpix2pix}, we adapt CFG to joint audio-video models with two conditioning sources: the textual prompt $c_T$ and the cross-modal information induced by Bridge interactions $c_B$. We derive a principled dual CFG formulation by decomposing the joint posterior via Bayes' rule:
\begin{equation}
P(z \mid c_T, c_B) = \frac{P(c_T \mid c_B, z) P(c_B \mid z) P(z)}{P(c_T, c_B)}.
\end{equation}
Taking the score function (gradient of log-likelihood) and applying CFG scaling leads to:
\begin{equation}
\begin{aligned}
\tilde{v}_\theta &= v_\theta(z_t, \varnothing, \varnothing) \\
&\quad + s_B \cdot \left[ v_\theta(z_t, \varnothing, c_B) - v_\theta(z_t, \varnothing, \varnothing) \right] \\
&\quad + s_T \cdot \left[ v_\theta(z_t, c_T, c_B) - v_\theta(z_t, \varnothing, c_B) \right],
\end{aligned}
\end{equation}

where $\varnothing$ denotes null conditioning, $v_\theta(z_t, c_T, c_B)$ is the model's velocity prediction with both text and Bridge active, and $v_\theta(z_t, \varnothing, \varnothing)$ disables both (the ``no-bridge'' mode disables cross-modal injection).

By tuning $s_B$ and $s_T$, we control the trade-off between alignment and perceptual quality. In the most general setting, we use \textbf{Dual CFG} with three function evaluations per step (NFE$=3$), and the two common special cases below reduce to NFE$=2$:
\begin{itemize}
    \item \textbf{Dual CFG (general)} ($s_B=s_b,\, s_T=s_t$): The full formulation that independently scales (i) modality-alignment guidance through the Bridge term and (ii) text guidance. This provides the most flexible control over the alignment--quality trade-off, at the cost of one additional model call (NFE$=3$).
    \item \textbf{Text-only CFG} ($s_B=1,\, s_T=s$): Standard formulation. Bridge remains active in both branches, so guidance does not explicitly amplify cross-modal alignment. Yields high semantic fidelity (e.g., ImageBind scores) but weaker temporal sync (higher DeSync). This is a two-branch guidance scheme (NFE$=2$).
    \item \textbf{Text + modality CFG} ($s_B=s,\, s_T=s$): The unconditional branch disables Bridge injection, isolating the alignment signal. Produces stronger synchronization (lower DeSync, better lip-sync). This also uses two branches (NFE$=2$).
\end{itemize}

\begin{figure}
  \centering
  \includegraphics[width=0.8\linewidth]{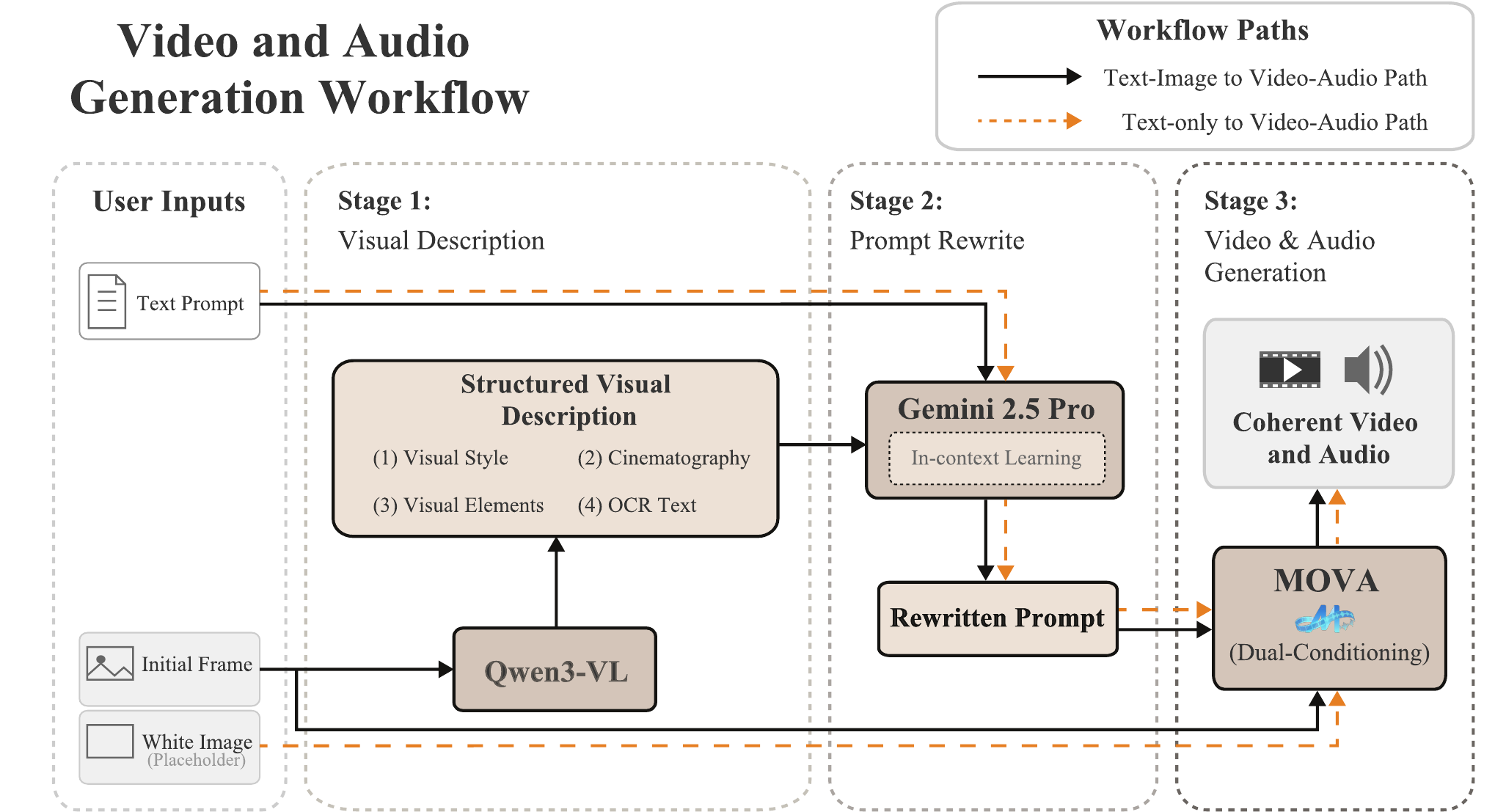}
  \caption{
  The overall workflow of MOVA for text-image and text-only to video-audio generation.}
  \label{fig:workflow}
\end{figure}

\subsection{Generation Workflow} 
\label{subsec:workflow}
To address diverse user input formats and maintain style consistency with training data, which, in our opinion, can incentivize model's full potential. Given a user-provided initial frame and a text prompt, our workflow generates coherent video and audio, as illustrated in Figure~\ref{fig:workflow}. The primary objective of this pipeline is prompt enhancement: rather than directly using raw user inputs which often lack descriptive detail, we refine them to incentivize the model’s full generative potential. We observe that maintaining high performance is closely tied to the quality of the prompt; therefore, our multi-stage conditioning pipeline explicitly extracts visual grounding from the reference image and synthesizes an enriched narrative. This ensures that the final synthesis not only preserves the style, lighting, and cinematography established in the first frame but also aligns with the sophisticated data distribution the model was trained on.

Our pipeline consists of three primary stages. First, we utilize Qwen3-VL~\cite{qwen3-vl}, a vision-language model, to extract a structured visual description. This extraction is guided by a curated prompt that constrains the model to four essential categories: (i) \textit{visual style}, including color palette and lighting; (ii) \textit{cinematography}, covering shot size, framing, and composition; (iii) \textit{visual elements}, comprising subjects and their spatial relations; and (iv) \textit{OCR text}, preserved exactly as it appears. This structured representation serves as an intermediate grounding that bridges the gap between the static image and the dynamic video.

Second, we synthesize a video generation prompt using LLMs (e.g. Gemini 2.5 Pro~\cite{gemini2_5_report_2025}, conditioned on both the input text and the extracted visual description. The core design principle is to preserve the static attributes from the visual description while incorporating the temporal dynamics specified by the user text. We employ in-context learning~\cite{in-context_learning} to ensure the generated prompt follows the narrative style and architecture of the training data.

Finally, MOVA generates video content by leveraging both the synthesized prompt and the initial frame as dual conditioning. This mechanism integrates narrative descriptions with visual grounding to ensure temporal synchronization while adhering to established visual priors. Furthermore, the versatility of our workflow allows for text-to-video-audio generation using a text prompt and an uninformative white image as input, thereby demonstrating MOVA's capacity for high-quality, zero-shot video synthesis.

% ===== Ablation Studies =====
% \section{Ablation Studies}
% \label{sec:ablations}

% \input{chapters/Ablations}

% ===== Evaluation =====
\section{Evaluation}
\label{sec:exp}

\subsection{Experiment Setup}
\label{sec:bench}

\paragraph*{Benchmarks.}
In this work, we use two benchmarks to test the model's video generation capabilities. We adopt Verse-Bench~\cite{wang2025universe}, which consists of 600 image--text prompt pairs. To facilitate joint video--audio generation, we employ GPT-5~\cite{chatgpt5} to unify visual and audio descriptions into a single, cohesive prompt.
While Verse-Bench~\cite{wang2025universe} provides a large-scale collection of image--text prompts, it is not specifically designed for evaluating video--audio generation in various scenes.
To further evaluate joint video--audio generation in realistic and challenging settings, we construct a dedicated evaluation benchmark covering diverse video generation scenarios. The benchmark is designed to assess capabilities that are critical for joint video--audio modeling, including temporal coherence, audio--visual synchronization, multi-character interaction, and dynamic scene evolution. Unlike Verse-Bench, which provides broad visual coverage, our benchmark adopts a finer-grained scenario taxonomy and explicitly categorizes samples into six representative video generation types, including multi-speaker interaction, movie-style narratives, sports competitions, game livestreams, camera motion sequences, and anime-style content. Detailed benchmark construction procedures and category descriptions are provided in Appendix~\ref{app:benchmark_details}.

\begin{figure}
  \centering
  \includegraphics[width=0.95\linewidth]{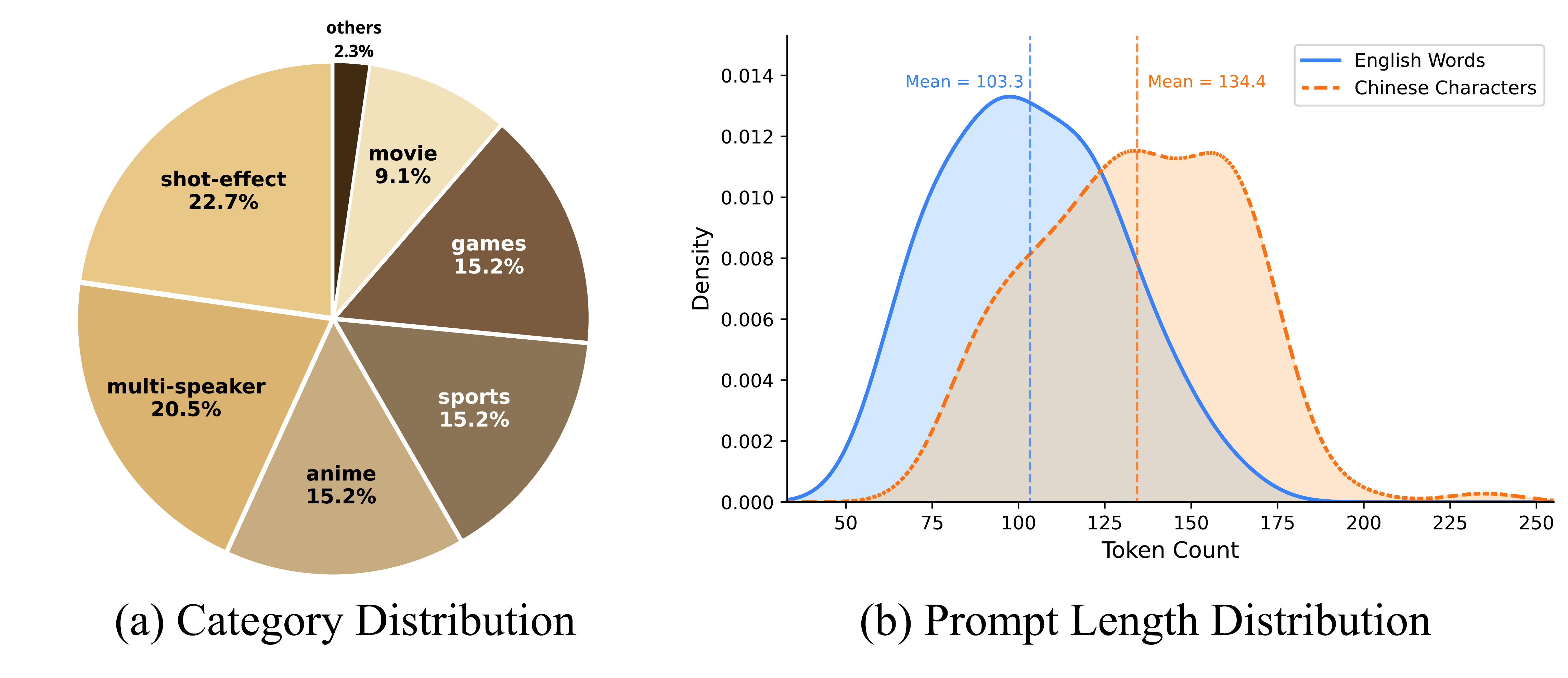}
  \caption{
  Dataset overview.
  (a) Category distribution of samples in the dataset, illustrating the relative proportions of different image categories.
  (b) Statistical distribution of prompt lengths.
  }
  \label{fig:benchmark_count}
\end{figure}

% \paragraph*{Setup}
% We first run each model on the 132 prompts to generate paired video and audio per prompt under a fixed inference configuration (frame rate, resolution, audio sample rate, and duration), and then compute objective metrics on the resulting set. See \Cref{sup:sec:details_of_exp} for more details about the settings.

% \paragraph*{Hardware setup}
% \cqy{It seems this paragraph can be removed.} For the \textbf{training stage}, we utilized computational resources equivalent to 1024 NVIDIA H200 GPUs, ensuring sufficient capacity to accommodate large-scale multimodal optimization. During \textbf{inference}, the evaluation was conducted with a significantly lighter configuration, relying on resources equivalent to a single NVIDIA H200 GPU. This setup highlights the efficiency of the proposed framework, as it requires massive parallel compute for training but maintains practical feasibility for inference on a single high-end GPU. \pxy{H200 GPU inference is not efficient at all. Simply remove this paragraph.}

\paragraph*{Evaluation metrics.}
We evaluate our method through both objective benchmarks and subjective human evaluations. 

\begin{itemize}
    \item \textbf{Objective evaluation}:
    For objective evaluation, we report performance on Verse-Bench across several dimensions. We measure acoustic fidelity and diversity using the Inception Score (IS) computed with a PANNs~\cite{panns} classifier, and assess speech quality with DNSMOS~\cite{dnsmos}. Cross-modal semantic alignment is quantified using the ImageBind score (IB-Score)~\citep{girdhar2023imagebind} computed on joint audio-video embeddings. Temporal consistency is measured by the DeSync score predicted by Synchformer~\cite{iashin2024synchformer}. Furthermore, to rigorously evaluate lip-sync precision, we include the Lip Sync Error - Confidence (LSE-C) and Distance (LSE-D) metrics derived from SyncNet~\cite{chung2016syncnet}. Additionally, we evaluate the concatenated minimum permutation character error rate (cpCER) score on the multi-speaker subset of our constructed benchmark. This metric is designed to assess whether the speaker identities and speaking content are correctly reflected in the generated outputs. To evaluate speaker timbre and dialogue content, we employ the MOSS Transcribe Diarize~\cite{yu2026moss}. This model performs speaker diarization using explicit speaker tags such as [S01] and [S02], followed by automatic speech recognition that transcribes the corresponding spoken content for each identified speaker.
    
    \item \textbf{Subjective Evaluation}: We conduct an arena-style preference study to evaluate human perception. The evaluation set comprises 732 samples, including 600 from Verse-Bench and 132 from our newly introduced benchmark, where half of the originally English-only Verse-Bench speech data was manually translated to construct a bilingual mix. For each comparison, participants are tasked with selecting the superior video across five dimensions: (i) \textit{prompt adherence}, (ii) \textit{visual-audio synchrony}, (iii) \textit{lip-sync accuracy}, (iv) \textit{video quality}, and (v) \textit{audio-speech fidelity}. A standard ELO rating system is employed to compute model rankings based on pairwise human preference judgments. Following the official Chatbot Arena implementation~\footnote{\url{https://colab.research.google.com/drive/1RAWb22-PFNI-X1gPVzc927SGUdfr6nsR}}, we set the initial ELO rating to 1000 with a K-factor of 4. The logistic scale and base are configured at 400 and 10, respectively. To ensure statistical robustness, we report the results using 1000 bootstrap iterations.
\end{itemize}

\paragraph*{Baselines.}
We compare MOVA with three baseline models under a unified protocol. Specifically, the baselines span two paradigms: (i) synchronous audio–visual generators that produce video and speech jointly: LTX-2~\cite{ltx-2} and Ovi~\cite{low2025ovi}. (ii) a cascaded pipeline formed by coupling WAN2.1~\cite{wan} for video with MMAudio~\citep{cheng2025mmaudio} for audio. For fair comparison, we standardize the spatial resolution at 720p and adopt the recommended configurations for all baseline models.

\begin{itemize}
   \item \textbf{Ovi}~\cite{low2025ovi} is a single-stage audio–video generation model that jointly models both audio and video modalities within a single process. By employing two architecturally DiTs for audio and video, Ovi achieves natural audiovisual synchronization without relying on separate pipelines or post hoc alignment. In the fusion blocks, OVI uses a frozen T5 encoder to integrate the video model with a pretrained audio model capable of generating both speech and environmental sounds. This shared semantic conditioning allows the audio and video branches to be guided by the same semantic context, strengthening cross-modal coherence and synchronization.
   
    \item \textbf{LTX-2}~\cite{ltx-2} introduces an efficient audiovisual generation model that, like OVI, produces video and its synchronized audio within a single diffusion process. Unlike OVI, LTX-2 adopts an asymmetric dual-stream Transformer architecture, enabling temporal alignment through bidirectional attention across all layers. The model employs modality-specific VAEs and positional encodings for audio and video, which preserve generation quality while significantly improving computational efficiency.
    
    \item \textbf{WAN2.1 + MMAudio} is a cascaded baseline that decouples video and audio generation. Specifically, we use WAN2.1~\cite{wan} as the video backbone, followed by MMAudio~\cite{cheng2025mmaudio} for video-conditioned audio synthesis. Unlike joint or synchronous modeling, this pipeline allocates the primary modeling burden to a strong video generator to capture spatiotemporal dynamics and kinematics, while the conditional audio generator focuses on acoustic consistency. This approach provides a competitive system-level baseline for audio-visual pairing.

\end{itemize}

% \begin{table}[t]
%     \centering
%     \caption{TO BE Replace}
%     \begin{tabular}{ccccccc}
%         \toprule
%          Dataset & Method & FVD ↓ & FAD ↓ & IB-AV ↑ \\
%         \midrule
%         \multirow{2}*{Landscape} & MM-Diffusion & 447 & 5.78 & 0.156 \\
%         & + MMDisCo & \textbf{405} & \textbf{5.52} & \textbf{0.162} \\
%         \midrule
%         \multirow{2}*{AIST++} & MM-Diffusion & 513 & 2.31 & 0.0897 \\
%         & + MMDisCo & \textbf{450} & \textbf{2.17} & \textbf{0.0909}\\
%         \bottomrule
%     \end{tabular}
%     \label{tab:in_domain}
% \end{table}
\subsection{Experimental Results}
\begin{table}[t]
  \centering
  \caption{Quantitative comparison of audio-visual generation performance on Verse-Bench. IS and AV-Align metrics are evaluated on all Verse-Bench subsets; DNSMOS and Lip Sync metrics are evaluated on Verse-Bench \textit{set3}; ASR Acc is evaluated on the \textit{multi-speaker} subset. \textbf{Bold} and \underline{underlined} values denote the best and second-best results, respectively.}
  \label{tab:baseline_comparison_selected}
  \setlength{\tabcolsep}{5pt}
  \renewcommand{\arraystretch}{1.2}
  \begin{tabular}{lcccccccc}
      \toprule
      \multirow{2}{*}{\textbf{Model}} & \multirow{2}{*}{$s_{B}$} & \multicolumn{2}{c}{\textbf{Audio-Speech}} & \multicolumn{2}{c}{\textbf{AV-Align}} & \multicolumn{2}{c}{\textbf{Lip Sync}} & \textbf{ASR Acc} \\
      \cmidrule(lr){3-4} \cmidrule(lr){5-6} \cmidrule(lr){7-8} \cmidrule(lr){9-9}
      & & IS$\uparrow$ & DNSMOS$\uparrow$ & DeSync$\downarrow$ & IB-Score$\uparrow$ & LSE-D$\downarrow$ & LSE-C$\uparrow$ & cpCER$\downarrow$ \\
      \midrule
      LTX-2 \cite{ltx-2}         & - & 3.066 & 3.635 & 0.451 & 0.213 & 7.261 & 6.109 & 0.220 \\
      Ovi \cite{low2025ovi}      & - & 3.680 & 3.516 & 0.515 & 0.190 & 7.468 & 6.378 & 0.436 \\
      WAN2.1 + MMAudio           & - &  4.036 & -- & \textbf{0.260} & \textbf{0.317} & -- & -- & -- \\
      \midrule
      MOVA-360p                  & 1.0 & \textbf{4.269} & \textbf{3.797} & 0.475 & 0.286 & 8.098 & 6.278 & \underline{0.177} \\
      \quad w/ dual CFG        & 3.5 & \underline{4.169}    & 3.674 & \underline{0.351}  & \underline{0.315}    & \textbf{7.004}  & \textbf{7.800}  &   0.247 \\
      MOVA-720p             & 1.0 & 3.936             & 3.671             & 0.485             & 0.277             & 8.048 & 6.593   &   \textbf{0.149} \\
      \quad w/ dual CFG    & 3.5 & 3.814 & \underline{3.751} & 0.370 & 0.297 & \underline{7.094} & \underline{7.452} & 0.218   \\
      \bottomrule
  \end{tabular}
\end{table}

\subsubsection{Comparison with Baseline Methods}
We evaluate the performance of MOVA against several competitive baselines, including LTX-2~\cite{ltx-2}, Ovi~\cite{low2025ovi}, and a cascaded pipeline (WAN2.1 + MMAudio). Table~\ref{tab:baseline_comparison_selected} presents a comprehensive quantitative comparison across four critical dimensions.

\paragraph{Audio Fidelity and Speech Quality.} 
MOVA demonstrates a clear advantage in generating high-quality audio. Specifically, MOVA-360p achieves a state-of-the-art IS of 4.269 and a DNSMOS~\cite{dnsmos} of 3.797, significantly outperforming LTX-2 and Ovi. We observe that while the cascaded WAN2.1 + MMAudio baseline shows respectable IS (4.036), it lacks the capability to generate intelligible speech content. In contrast, MOVA maintains high speech naturalness and clarity even as we scale the resolution to 720p (DNSMOS of 3.671), suggesting that our unified modeling effectively captures the complex distribution of human speech and ambient sounds.

\paragraph{Audio-Visual Alignment.} 
We evaluate audio-visual alignment through two lenses: temporal synchronization (DeSync~\cite{iashin2024synchformer}) and cross-modal semantic alignment (IB-Score~\cite{girdhar2023imagebind}). Compared to contemporary unified models like LTX-2 and Ovi, MOVA exhibits a pronounced advantage. Specifically, MOVA-360p with dual CFG ($s_{B}=3.5$) achieves a DeSync of 0.351 and an IB-Score of 0.315, significantly surpassing LTX-2 (0.451 / 0.213) and Ovi (0.515 / 0.190). This substantial margin, particularly the $\sim$50\% improvement in IB-Score over Ovi, suggests that MOVA effectively binds auditory events to generated visual context. Notably, although the specialized cascaded pipeline (WAN2.1 + MMAudio) yields the best DeSync (0.260) due to its task-specific audio generator, MOVA virtually closes the gap in both temporal and semantic metrics. This demonstrates that our dual CFG strategy effectively amplifies the cross-modal alignment signal during inference, allowing a unified architecture to match the precision of modularized pipelines without sacrificing structural simplicity or end-to-end coherence.

\paragraph{Lip-Sync Precision.} 
The accuracy of fine-grained lip synchronization is measured by LSE-D and LSE-C~\cite{chung2016syncnet}. MOVA variants, particularly those equipped with dual CFG, exhibit a dominant performance in this category. MOVA-360p w/ dual CFG achieves the best LSE-D (7.004) and LSE-C (7.800), representing a substantial margin over LTX-2 ($7.261/6.109$) and Ovi ($7.468/6.378$). Interestingly, even without dual CFG, MOVA-720p maintains a competitive LSE-C of 6.593. This superior lip-sync performance confirms that our model has internalized a sophisticated mapping between phonetic features and labial dynamics, which is further refined by the explicit guidance of the Bridge interactions during inference.

\paragraph{Multi-Speaker Attribution.}
As illustrated in Table~\ref{tab:baseline_comparison_selected}, MOVA-720p achieves the cpCER of 0.149, lower than LTX-2 (0.220) and Ovi (0.436). The high error rate of Ovi in this metric suggests a frequent "voice-identity mismatch," where the model fails to associate the correct speech with the corresponding subject. The low cpCER of MOVA demonstrates its ability to maintain high-fidelity identity consistency and correct audio-visual attribution in crowded scenes, a crucial requirement for realistic content generation. For both MOVA-360p and MOVA-720p, the introduction of dual CFG leads to a slight increase in cpCER. This is primarily because dual CFG redistributes the guidance strength among multiple conditional branches during sampling, thereby diluting the relative weight of the text condition and reducing the model’s ability to follow explicit speaker instructions, such as speaker tags [S01]/[S02] and their corresponding speech content. In multi-speaker scenarios, this weakened instruction-following behavior more easily results in mismatches between speaker identities and transcribed content, which is reflected in higher cpCER. In addition, compared to MOVA-360p, MOVA-720p consistently achieves lower cpCER under both settings. This advantage mainly stems from differences in training data scale and training stages. MOVA-720p is further fine-tuned on top of MOVA-360p with broader data coverage and more extensive exposure to multi-speaker samples, leading to improved stability in speaker identity and speech content association.

\begin{figure}
  \centering
  \includegraphics[width=0.55\linewidth]{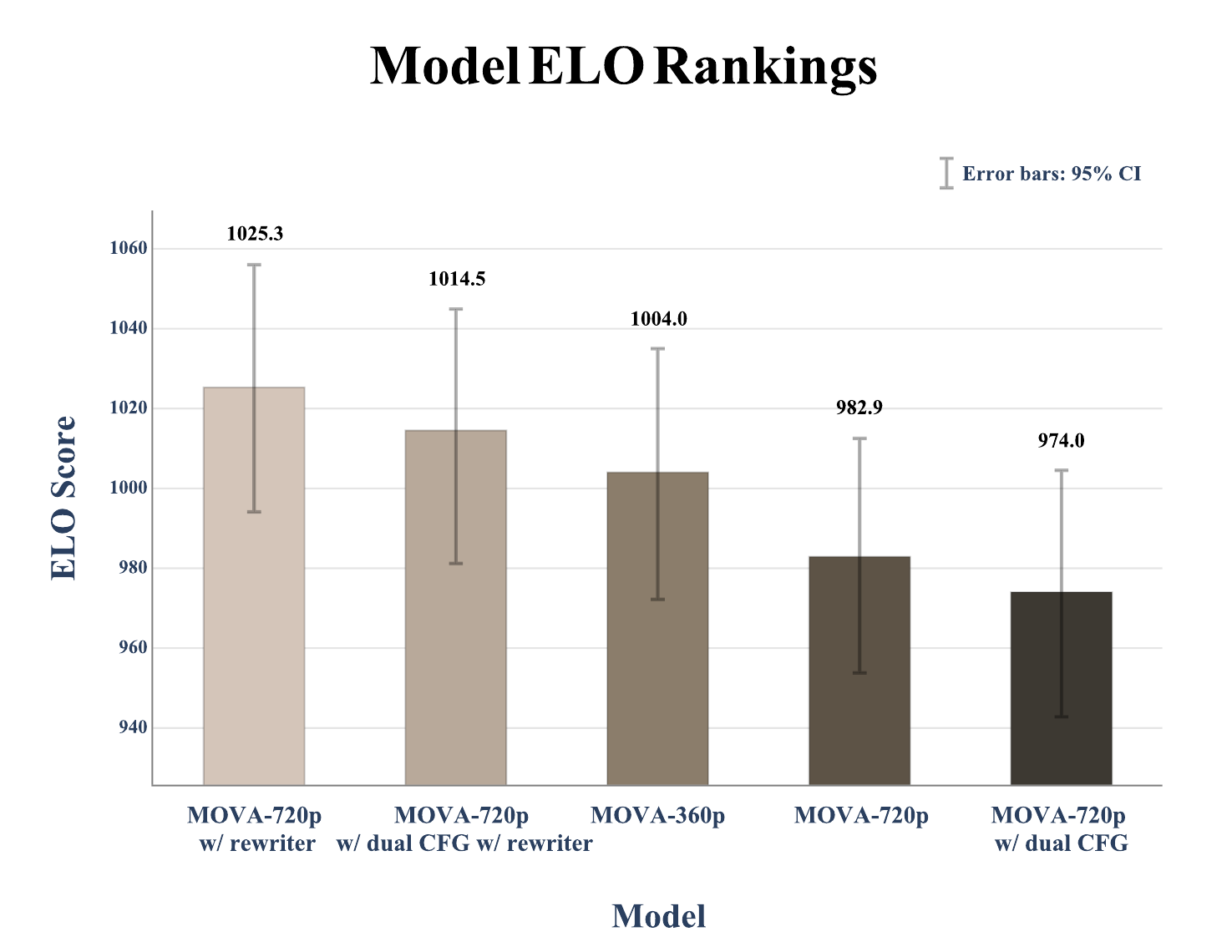}
  \caption{
  Ablation study on human preference.
  }
  \label{fig:arena_results_abla}
\end{figure}

\subsubsection{Ablation Study}
We conduct a series of ablation experiments of MOVA to evaluate the impact of our training strategy and inference configurations, which is shown in Table~\ref{tab:baseline_comparison_selected}, Table~\ref{tab:ablation_of_cfg} and Table~\ref{tab:ablation}.

\paragraph{Scaling to High Resolution.} 
We evaluate the empirical robustness of MOVA by scaling the generation resolution from 360p to 720p. As summarized in Table~\ref{tab:baseline_comparison_selected}, the 720p variant demonstrates remarkawble consistency across diverse evaluation dimensions. Specifically, in terms of temporal and semantic alignment, MOVA-720p maintains a DeSync of 0.485 and an IB-Score of 0.277, showing negligible degradation compared to the 360p base model (0.475 and 0.286, respectively). This stability is particularly noteworthy as increasing visual resolution often introduces challenges in maintaining cross-modal coherence. Additionally, MOVA-720p achieves a Lip Sync LSE-C of 6.593, outperforming the 360p version’s 6.278. Similarly, it reports the best-ever cpCER (0.149) on the multi-speaker subset. While there is a marginal decrease in audio fidelity and speech quality metrics, which is a common trade-off when the model capacity is further distributed to handle increased visual complexity, the overall performance profile remains highly competitive. These results collectively validate our staged training strategy, confirming that MOVA can effectively scale to high-resolution synthesis while preserving its foundational audio-visual generative priors.

\begin{table}[t]
    \centering
    \caption{Ablation study of $s_{B}$. IS and AV-Align metrics are evaluated on all Verse-Bench subsets; DNSMOS and Lip Sync metrics are evaluated on Verse-Bench \textit{set3}; ASR Acc is evaluated on the \textit{multi-speaker} subset. \textbf{Bold} and \underline{underlined} values denote the best and second-best results, respectively.}
    \label{tab:ablation_of_cfg}
    \setlength{\tabcolsep}{6pt}
    \renewcommand{\arraystretch}{1.2}
    \begin{tabular}{lcccccccc}
        \toprule
        \multirow{2}{*}{\textbf{Model}} & \multirow{2}{*}{\textbf{$s_{B}$}} & \multicolumn{2}{c}{\textbf{Audio-Speech}} & \multicolumn{2}{c}{\textbf{AV-Align}} & \multicolumn{2}{c}{\textbf{Lip Sync}} & \textbf{ASR Acc} \\
      \cmidrule(lr){3-4} \cmidrule(lr){5-6} \cmidrule(lr){7-8} \cmidrule(lr){9-9}
         & & IS$\uparrow$ & DNSMOS$\uparrow$ & DeSync$\downarrow$ & IB-Score$\uparrow$ & LSE-D$\downarrow$ & LSE-C$\uparrow$ & cpCER$\downarrow$\\
        \midrule
        MOVA-360p           & 1.0  &   \underline{4.269}   & \textbf{3.797}    & 0.475             & 0.286 & 8.098             & 6.278 & \textbf{0.177}\\
        \quad w/ dual CFG        & 2.0 & 4.222    & \underline{3.748} & 0.421 & 0.305             & 7.323             & 7.331 & \underline{0.185} \\
                & 3.0 & \textbf{4.319}    & 3.686 & 0.388 & 0.312             & 7.014             & 7.774 & 0.188 \\
                & 3.5 & 4.169    & 3.674 & \textbf{0.351}  & \underline{0.315}    & \underline{7.004}  & \underline{7.800}  &   0.247  \\
                & 4.0 & 4.225    & 3.631 & \underline{0.365} & \textbf{0.316}             & \textbf{6.957}             & \textbf{7.891} & 0.264 \\
        \bottomrule
    \end{tabular}
\end{table}
\paragraph{Effect of Dual Classifier-Free Guidance.} 
We evaluate the influence of the dual CFG scale $s_B$ in Table \ref{tab:ablation_of_cfg}. Our results demonstrate a synergistic improvement across all alignment-related metrics as $s_B$ increases from 1.0 to 4.0. Specifically, we observe a consistent reduction in DeSync and LSE-D, alongside a significant gain in IB-Score and LSE-C. For instance, as $s_B$ scales to 4.0, the LSE-C reaches a peak of 7.891 and the DeSync score is minimized to 0.365. This uniform progression across multiple distinct metrics suggests that strengthening the modality-alignment guidance effectively improves the synchronization between generated video and audio. However, this heightened alignment precision comes at a clear cost to speech quality and instruction following. As the guidance toward audio-visual alignment becomes more dominant, we observe a concurrent degradation in DNSMOS and a rise in cpCER (from 0.177 to 0.264). We interpret this phenomenon as a form of conditional interference: in the multi-branch sampling process, excessively high $s_B$ prioritizes the geometric constraints of synchronization over the generative fidelity of the speech signal. This "over-regularization" potentially leads to a diminished sensitivity to textual instructions (e.g., speaker-specific tags), causing the model to prioritize how the speech aligns with the video at the expense of what is being said and how natural it sounds.

\paragraph{Emergent T2VA Capability.} 
Interestingly, we find that MOVA exhibits a strong emergent capability for the T2VA task, which is summarized in Table~\ref{tab:ablation}. By substituting the reference image with a null placeholder (MOVA-360p-T2VA), we test whether the model can synthesize synchronized content driven solely by textual prompts. As summarized in Table~\ref{tab:ablation}, several intriguing observations emerge. Notably, the T2VA variant achieves a superior IS of 4.370 and a lower DeSync of 0.441 compared to the standard TI2VA baseline. This performance gain suggests that in the absence of explicit structural constraints from a reference image, the model can more freely explore the joint audio-visual manifold, leading to higher audio fidelity and improved temporal synchronization. Predictably, identity-dependent metrics like LSE-C and LSE-D show a marginal decline, as the null placeholder provides no lip geometry. However, the overall stability of the T2VA results is remarkable. This underscores that MOVA has internalized a robust, decoupled yet highly coordinated prior for video and audio synthesis, allowing it to maintain temporal synchronization and semantic alignment even when the visual conditioning is entirely removed.

\subsubsection{Arena-Based Human Evaluation}
Given the limitations of current objective evaluation frameworks for audio-visual generation models, MOVA introduces an Arena-based human preference evaluation paradigm that includes the latest open-source audio-visual generation models worldwide. The evaluation collected over 5,000 valid votes and systematically analyzed the results. To ensure a fair comparison, all models utilize their respective official prompt-refinement methods (e.g., our rewriter described in Section~\ref{subsec:workflow}) to enhance the video generation prompt.

\begin{figure}[t]
    \centering
    \begin{subfigure}[b]{0.4\textwidth}
        \centering
        \includegraphics[width=\textwidth]{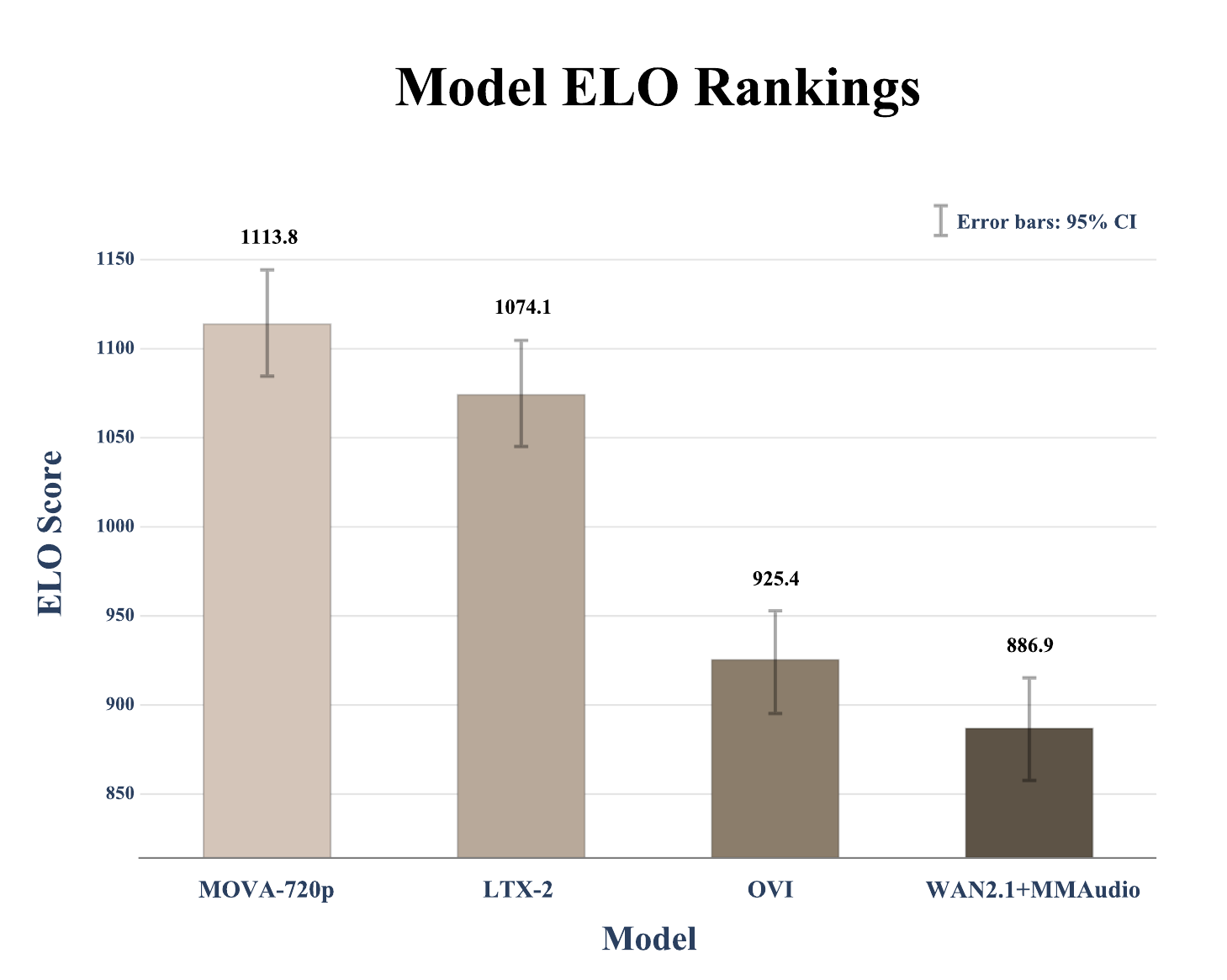}
        \caption{Human preference ELO ranking.}
        \label{fig:elo_ranking}
    \end{subfigure}
    \hfill
    \begin{subfigure}[b]{0.55\textwidth}
        \centering
        \includegraphics[width=\textwidth]{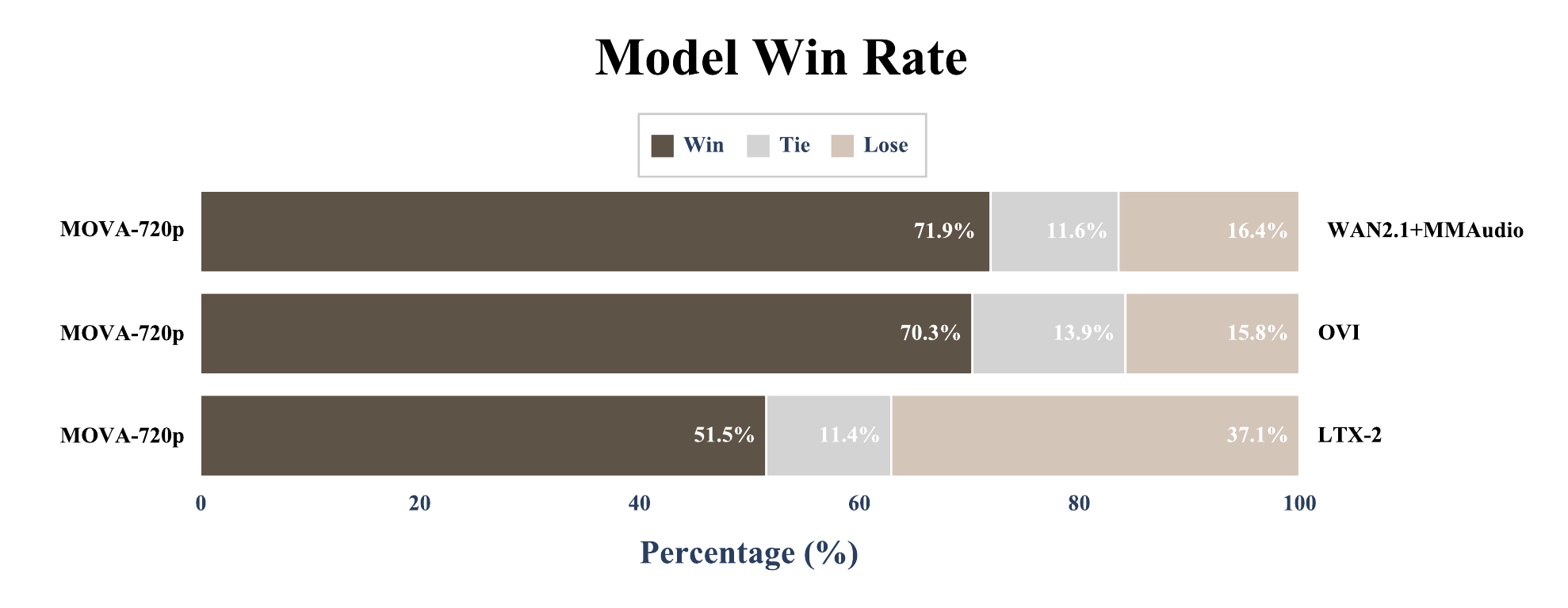}
        \caption{Win rates of MOVA against baseline models.}
        \label{fig:win_rate}
    \end{subfigure}
    \caption{Arena evaluation results showing MOVA's performance in human preference studies. (a) ELO ratings demonstrate MOVA's superiority over baseline models. (b) Pairwise win rates show MOVA consistently outperforms all competitors, particularly against Ovi and the WAN + MMAudio cascade.}
    \label{fig:arena_results}
\end{figure}

\paragraph{Subjective Comparison with Baseline Methods.} As shown in Figure~\ref{fig:arena_results}, MOVA demonstrates a clear superiority in human preference: it is more frequently selected by users in pairwise comparisons, achieving an ELO rating of 1113.8 (starting from an initial rating of 1000), significantly higher than all baseline models. MOVA consistently maintains a win rate exceeding 50\%, with win rates surpassing 70\% against Ovi and the cascaded system (WAN + MMAudio).

\paragraph{Subjective Ablation Study.} As illustrated in Figure~\ref{fig:arena_results_abla}, we conduct an internal Arena to dissect the impact of prompt refinement, resolution scaling, and inference strategies on human preference. A primary observation is the critical role of the prompt rewriter. Variants utilizing refined prompts consistently achieve superior ELO ratings, with MOVA-720p (w/ rewriter) reaching the peak score of 1025.3. Compared to the standard MOVA-720p (982.9), this substantial ELO gain validates our motivation: user-provided inputs often vary in format and level of detail, creating a distribution gap with the model's training data. By employing our multi-stage conditioning pipeline, which leverages LLMs to synthesize prompts that preserve visual grounding (e.g., style, cinematography) while incorporating temporal dynamics, we bridge this gap and effectively incentivize the model's full generative potential. Regarding our inference strategy, we observe a subtle trade-off; while dual CFG ($s_{B}=3.5$) significantly improves objective alignment metrics, it leads to a slight decrease in human preference scores, dropping from 1025.3 to 1014.5 in the rewriter-enhanced 720p models. We attribute this decrease to the formulation of dual CFG: by explicitly amplifying cross-modal alignment signals, the relative guidance scale for the primary text instruction is effectively diminished. This can occasionally result in reduced instruction-following. 
\begin{table}
    \centering
    \caption{Evaluation of T2VA effectiveness. IS and AV-Align metrics are evaluated on all Verse-Bench subsets; DNSMOS and Lip Sync metrics are evaluated on Verse-Bench \textit{set3}; ASR Acc is evaluated on the \textit{multi-speaker} subset. \textbf{Bold} values denote the best results.}
    \label{tab:ablation}
    \setlength{\tabcolsep}{6pt}
    \renewcommand{\arraystretch}{1.2}
    \begin{tabular}{lccccccc}
        \toprule
        \multirow{2}{*}{\textbf{Model}} & \multicolumn{2}{c}{\textbf{Audio-Speech}} & \multicolumn{2}{c}{\textbf{AV-Align}} & \multicolumn{2}{c}{\textbf{Lip Sync}} & \textbf{ASR Acc} \\
      \cmidrule(lr){2-3} \cmidrule(lr){4-5} \cmidrule(lr){6-7} \cmidrule(lr){8-8}
        & IS$\uparrow$ & DNSMOS$\uparrow$ & DeSync$\downarrow$ & IB-Score$\uparrow$ & LSE-D$\downarrow$ & LSE-C$\uparrow$ & cpCER$\downarrow$\\
        \midrule
        MOVA-360p             &   4.269   & \textbf{3.797}    & 0.475             & \textbf{0.286} & \textbf{8.098}             & \textbf{6.278} & \textbf{0.177}\\
        MOVA-360p-T2VA        & \textbf{4.370}    & 3.767 & \textbf{0.441} & 0.281             & 8.362             & 5.830 & 0.188 \\
        \bottomrule
    \end{tabular}
\end{table}

\subsection{Scaling to Lip Synchronization}
\label{sec:scaling_lip}
Lip synchronization is among the most demanding audio--video generation tasks. Unlike discrete, onset-driven events (e.g., ``chopping fruit'' or ``hitting a drum'') where alignment depends on a few salient temporal onsets, speech requires continuous, fine-grained correspondence between mouth shapes and phonemes across long spans. We find that architectural mechanisms alone (e.g., Bridge modules for cross-modal attention) are insufficient to achieve high-quality lip synchronization---the model must also learn phoneme-to-viseme mappings from data, which requires larger capacity and more training examples.

\begin{figure}[t]
    \centering
    \includegraphics[width=0.8\linewidth]{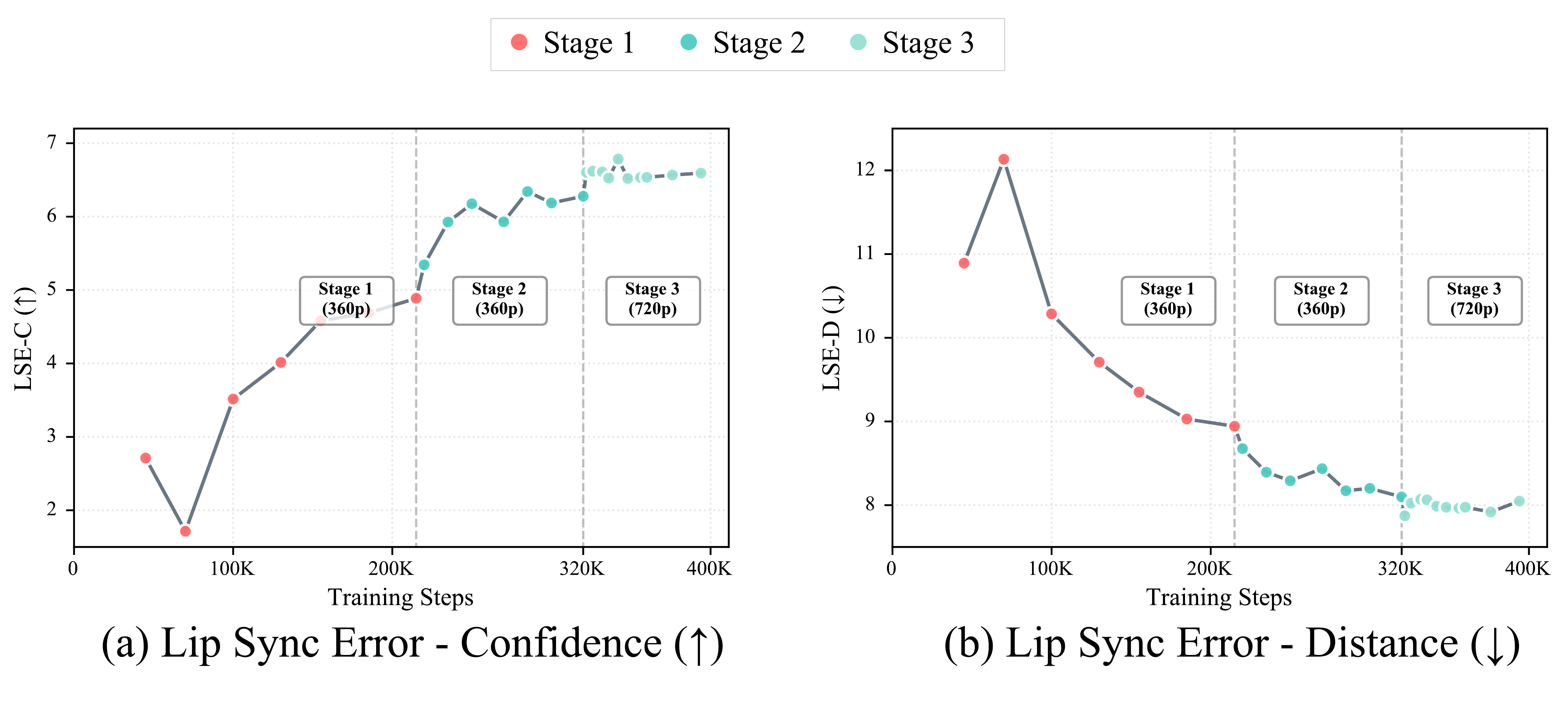}
    \caption{Training progression across three stages. Stage 1 (360p, aggressive bridging) establishes basic alignment; Stage 2 (360p, aligned schedules) refines consistency; Stage 3 (720p) scales to high resolution while preserving synchronization.}
    \label{fig:training_curves}
\end{figure}

Figure~\ref{fig:training_curves} shows the progression of LSE-C (higher is better) and LSE-D (lower is better) across our three-stage training process. In Stage 1, we train at 360p with aggressive video denoising, mild audio denoising, and high text dropout to force the model to rely on cross-modal bridging for alignment. LSE-D drops rapidly and LSE-C rises, indicating the model quickly learns basic synchronization patterns. Stage 2 maintains 360p but aligns the noise schedules across modalities for more stable cross-modal attention, reduces text dropout to refine semantic details, and applies loudness normalization to avoid CFG-induced volume distortion. LSE-D continues to decrease while LSE-C shows a notable jump, reflecting improved consistency and confidence in alignment. Finally, Stage 3 scales to 720p. With stable cross-modal alignment already established, the model can safely allocate capacity to higher resolution and finer spatial details without disrupting the learned synchronization structure. LSE-D further decreases and plateaus, while LSE-C stabilizes at a high level, indicating convergence to high-quality lip synchronization.

% ===== Discussion =====
\section{Discussion}
\label{sec:discussion}

\subsection{Predefined sigma as an Implicit Synchronization Prior} 

Audio-visual synchronization inherently contains a tension in conditioning direction. For many discrete, event-driven cases, the most natural formulation is Video$\rightarrow$Audio: the visual stream deterministically anchors when and where events occur, and most sound events (impacts, collisions, cuts, percussive gestures) are visually driven with clear temporal onsets. In these settings, generating audio conditioned on video aligns with both human intuition and the causal structure of the scene. In contrast, other synchronization tasks are more realistically Audio$\rightarrow$Video and better match practical applications. Speech-driven generation is the canonical example: given an audio track, the target is to produce temporally consistent facial motion and lip articulation (Speech$\rightarrow$Video), potentially with speaker-specific dynamics, language-dependent phoneme--viseme mappings, and context-dependent expressiveness. This directionality is often required in downstream pipelines such as dubbing, avatar animation, and talking-head generation, where audio is fixed and visuals must adapt.

This tension becomes more pronounced under the diffusion formulation. At each timestep, the noise levels for video and audio latents are governed by pre-defined schedules $\sigma_v(t_v)$ and $\sigma_a(t_a)$ (introduced in \nameref{par:dual-sigma-shift}), which act as fixed priors and are not learnable during training. As a result, the relative corruption of the audio stream is largely fixed by design, whereas the effective uncertainty in the visual stream can vary substantially with object scale and visual dominance in the frame (as discussed in Stable Diffusion 3~\citep{esser2024scaling}). Consequently, the same global schedule can implicitly bias the conditional direction: for close-up shots of lip motion where the target region occupies a large portion of the frame, the visual latent is relatively informative and the generation tends to behave like Video$\rightarrow$Audio; conversely, when the relevant speaker occupies only a small region, the visual evidence becomes comparatively uncertain, and the process may naturally shift toward Audio$\rightarrow$Video, letting speech provide the more reliable temporal anchor.

\subsection{Classifier-Free Guidance: factorization order}
% \paragraph{Discussion: why this factorization order.}

Our dual-CFG derivation starts from the factorization
$
P(z \mid c_T, c_B) \propto P(c_T \mid c_B, z)\, P(c_B \mid z)\, P(z),
$
which induces a natural nesting structure: we first ``turn on'' cross-modal information (via $c_B$) and then apply text guidance on top of it (via $c_T$). This order is not the only valid choice. For example, one may alternatively decompose the posterior as
\begin{equation}
P(z \mid c_T, c_B) \propto P(c_B \mid c_T, z)\, P(c_T \mid z)\, P(z),
\end{equation}
and derive an analogous two-term guidance by swapping the roles of $c_T$ and $c_B$.

We adopt our order for a practical reason: it admits a clean reduction to the standard text-only CFG as a special case. Concretely, setting $s_B=1$ makes the Bridge condition appear in both the conditional and ``unconditional'' text branches, so the guidance collapses to the familiar text-CFG form while keeping cross-modal injection fixed. In other words, by tuning $(s_B, s_T)$ we can continuously interpolate between ``text-only CFG'' and ``text + modality CFG'' without changing the sampling procedure.

In contrast, under the alternative factorization, the branch structure couples $c_B$ to the text-unconditional baseline in a way that prevents an equivalent ``text-only CFG'' limit: there is no choice of guidance weights that keeps the Bridge behavior identical across the two text branches while still producing the standard text-CFG difference term. As a result, the swapped order does not provide an intuitive knob for ``only amplify text while leaving cross-modal interactions unchanged.''

\subsection{Limitations}

\paragraph{Audio Modeling Capacity and Coverage.}
Our audio tower follows the Wan2.1-1.3B backbone, which may limit the modeling capacity for acoustically rich or highly structured signals. In particular, we observe degraded performance on singing voice, complex sound textures, and music/instrumental content, where fine-grained pitch/harmonic structure and long-range temporal dependencies are critical. More broadly, some audio--visual phenomena require stronger physical reasoning (e.g., correctly reflecting propagation delays such as the temporal offset between lightning and thunder), which is not explicitly enforced by our current objective and data.

\paragraph{Multi-speaker Synchronization and Annotation Reliability.}
While our model can handle single-speaker lip-sync reliably in many cases, multi-speaker scenes remain challenging. Rapid speaker turn-taking, overlapping speech, and ambiguous on-screen attribution can lead to incorrect mouth--audio assignment and temporal drift. This issue is compounded by the data pipeline: diarization errors and imperfect active-speaker labels can propagate to training, making the model conflate speakers or learn inconsistent supervision. Improving multi-speaker supervision (e.g., stronger active-speaker detection, cross-modal speaker tracking, and better filtering of noisy segments) is necessary for robust deployment.

\paragraph{Sequence Length and Computational Cost.}
Our current training and inference are constrained by sequence length. For example, a 720p, 8\,s clip at 24\,fps yields on the order of $1.6\times 10^5$ tokens, resulting in high memory and compute costs. This limits throughput during training and increases latency at inference, especially when using the most general guidance setting (NFE$=3$). Future work could address this bottleneck via more aggressive temporal/spatial compression, hierarchical or blockwise generation, and system-level optimizations tailored to long-context video tokens.

% ===== Related Work =====
\section{Related Work}
\label{sec:related}

\paragraph{Video Generation.}
Diffusion transformers (DiTs) \citep{ho2020denoising,DiT} have enabled large-scale video synthesis. Open models such as Wan \citep{wan} and HunyuanVideo \citep{kong2024hunyuanvideo} achieve near-photorealistic generation through efficient attention \citep{dao2022flashattention,dao2023flashattention2} and transformer scaling \citep{kaplan2020scaling}. Recent work extends these models to long-horizon generation \citep{longcatvideo2025}, controllable camera motion \citep{he2024cameractrl}, and high-resolution outputs exceeding 1080p. However, most text-to-video systems remain video-only, leaving audio generation as a separate problem. Proprietary systems Veo3 \citep{google-veo3} and Sora2 \citep{sora22025} demonstrate joint audio-video capabilities, but their closed-source nature limits reproducibility.

\paragraph{Audio Generation and Cascaded Pipelines.}
Latent diffusion enables scalable text-to-audio generation \citep{liu2023audioldm,liu2023audioldm2}. Audio VAEs compress waveforms into compact latent representations: DAC \citep{kumar2024high} uses residual vector quantization for high-quality reconstruction, while Stable Audio \citep{evans2025stable} employs a stereo variational autoencoder with spectral losses. Cascaded video-to-audio (V2A) pipelines \citep{luo2023difffoley,zhang2024foleycrafter,cheng2025mmaudio} represent a prevalent approach for audiovisual content creation. MMAudio \citep{cheng2025mmaudio} achieves temporal alignment by extracting features from video and using them as conditioning signals. While cascaded pipelines utilize strong single-modality priors, the sequential factorization ignores bidirectional modality influence: audio cannot inform visual trajectory during sampling, and vice versa.

\paragraph{Joint Audio-Video Generation.}
End-to-end joint generation has been explored to overcome cascaded limitations \citep{MTV,liu2025javisdit,ruan2022mmdiffusion,low2025ovi,wang2025universe,zhang2025uniavgen,hu2025harmony}. MMDisCo \citep{hayakawa2024mmdisco} uses discriminator-guided cooperative diffusion to align pretrained models, though adversarial training introduces instability at scale. MM-Diffusion \citep{ruan2022mmdiffusion} and JavisDiT \citep{liu2025javisdit} propose dual-stream architectures with cross-modal attention but are restricted to ambient sounds. MTV \citep{MTV} explicitly separates audio into speech, effects, and music tracks with disentangled control over lip motion, event timing, and visual mood, achieving precise audio-visual synchronization across diverse audio types. UniVerse-1 \citep{wang2025universe} integrates Wan2.1 and Ace through a stitching-of-experts paradigm with independent noise sampling, but suffers from audio-video drift. Recent works such as Ovi \citep{low2025ovi}, Harmony \citep{hu2025harmony}, and UniAVGen \citep{zhang2025uniavgen} adopt dual-tower architectures with RoPE-based positional encoding, achieving lip-synchronized video generation results without requiring prior separation of speech and environmental sounds. However, they have not scaled to general domains to demonstrate the full potential of the architecture. LTX-2 \citep{ltx-2} successfully scales the dual-tower approach to cover both lip-synchronized speech and general domain sounds through large-scale data training, though the audio quality exhibits some electronic artifacts that require cleaner reconstruction. We address these limitations through capacity scaling with a 29B dual-tower architecture, achieving high-fidelity audio and strong performance on both lip-synchronized speech and general domain sounds across bilingual settings.

% ===== Conclusions =====
\section{Conclusions}
\label{sec:conclusion}

We presented MOVA, an open and scalable framework for joint video--audio generation with 32B total parameters (18B active). MOVA uses an asymmetric dual-tower design (an A14B video backbone and a 1.3B audio backbone) together with a 2.6B bidirectional bridge and Aligned RoPE to support fine-grained temporal audio--visual interaction.

Our work targets three key challenges in joint audio--video generation: data, modeling, and scaling. We curate over 100,000 hours of fine-grained audio--visual data with sound, music, and speech annotations aligned to visual content. We also propose training and architectural designs that improve the stability of large-scale multimodal diffusion training, including decoupled timestep sampling to allow modality-specific noise schedules. Through controlled scaling studies, we find that increasing video model capacity and training data substantially improves lip synchronization, where smaller models show clear performance saturation.

In addition, we report practical system optimizations for large-scale training (Context Parallelism, FSDP2 strategies, and scheduled garbage collection), enabling stable 1024-GPU runs and achieving approximately 35\% MFU. We will release the model weights, training code, and inference pipelines, and we hope MOVA can serve as a strong open baseline for future research on synchronized audio--video generation. Finally, important limitations remain, including high training cost and remaining failure cases in motion generation, which we leave for future work.

% ===== Contributors =====
\newpage
\section*{Contributors}

\textbf{Core Contributors and Contributors are sorted alphabetically by first name, excluding advisors.}

\noindent\textbf{Core Contributors}: \\
Donghua Yu, Mingshu Chen, Qi Chen, Qi Luo, Qianyi Wu, Qinyuan Cheng\textsuperscript{*}, Ruixiao Li, Tianyi Liang\textsuperscript{*}, Wenbo Zhang, Wenming Tu, Xiangyu Peng, Yang Gao, Yanru Huo, Ying Zhu, Yinze Luo, Yiyang Zhang, Yuerong Song, Zhe Xu, Zhiyu Zhang

\vspace{0.5em}

\noindent\textbf{Contributors}: \\
Chenchen Yang, Cheng Chang, Chushu Zhou, Hanfu Chen, Hongnan Ma, Jiaxi Li, Jingqi Tong, Junxi Liu, Ke Chen, Shimin Li, Songlin Wang, Wei Jiang, Zhaoye Fei, Zhiyuan Ning

\vspace{0.5em}

\noindent\textbf{Advisors}: Chunguo Li, Chenhui Li, Ziwei He, Zengfeng Huang, Xie Chen$^\dagger$, Xipeng Qiu$^\dagger$

\vspace{1em}

\noindent\textbf{Affiliations}: \\
Shanghai Innovation Institute\\
MOSI Intelligence\\
Fudan University\\
Shanghai Jiao Tong University\\
East China Normal University\\
Tongji University\\
Southeast University\\
Xiamen University\\
University of Electronic Science and Technology of China

{\let\thefootnote\relax\footnotetext{$^*$Project Lead. $^\dagger$Corresponding authors: \texttt{chenxie95@sjtu.edu.cn}, \texttt{xpqiu@fudan.edu.cn}}}

% ===== Bibliography =====
\clearpage
\bibliographystyle{unsrtnat}
\bibliography{main}

@inproceedings{dnsmos,
  author       = {Chandan K. A. Reddy and
                  Vishak Gopal and
                  Ross Cutler},
  title        = {Dnsmos: {A} Non-Intrusive Perceptual Objective Speech Quality Metric
                  to Evaluate Noise Suppressors},
  booktitle    = {{IEEE} International Conference on Acoustics, Speech and Signal Processing,
                  {ICASSP} 2021, Toronto, ON, Canada, June 6-11, 2021},
  pages        = {6493--6497},
  publisher    = {{IEEE}},
  year         = {2021},
  url          = {https://doi.org/10.1109/ICASSP39728.2021.9414878},
  doi          = {10.1109/ICASSP39728.2021.9414878},
  bibsource    = {dblp computer science bibliography, https://dblp.org}
}

@article{panns,
  author       = {Qiuqiang Kong and
                  Yin Cao and
                  Turab Iqbal and
                  Yuxuan Wang and
                  Wenwu Wang and
                  Mark D. Plumbley},
  title        = {PANNs: Large-Scale Pretrained Audio Neural Networks for Audio Pattern
                  Recognition},
  journal      = {{IEEE} {ACM} Trans. Audio Speech Lang. Process.},
  volume       = {28},
  pages        = {2880--2894},
  year         = {2020},
  url          = {https://doi.org/10.1109/TASLP.2020.3030497},
  doi          = {10.1109/TASLP.2020.3030497},
  bibsource    = {dblp computer science bibliography, https://dblp.org}
}

@inproceedings{in-context_learning,
  author       = {Tom B. Brown and
                  Benjamin Mann and
                  Nick Ryder and
                  Melanie Subbiah and
                  Jared Kaplan and
                  Prafulla Dhariwal and
                  Arvind Neelakantan and
                  Pranav Shyam and
                  Girish Sastry and
                  Amanda Askell and
                  Sandhini Agarwal and
                  Ariel Herbert{-}Voss and
                  Gretchen Krueger and
                  Tom Henighan and
                  Rewon Child and
                  Aditya Ramesh and
                  Daniel M. Ziegler and
                  Jeffrey Wu and
                  Clemens Winter and
                  Christopher Hesse and
                  Mark Chen and
                  Eric Sigler and
                  Mateusz Litwin and
                  Scott Gray and
                  Benjamin Chess and
                  Jack Clark and
                  Christopher Berner and
                  Sam McCandlish and
                  Alec Radford and
                  Ilya Sutskever and
                  Dario Amodei},
  editor       = {Hugo Larochelle and
                  Marc'Aurelio Ranzato and
                  Raia Hadsell and
                  Maria{-}Florina Balcan and
                  Hsuan{-}Tien Lin},
  title        = {Language Models are Few-Shot Learners},
  booktitle    = {Advances in Neural Information Processing Systems 33: Annual Conference
                  on Neural Information Processing Systems 2020, NeurIPS 2020, December
                  6-12, 2020, virtual},
  year         = {2020},
  url          = {https://proceedings.neurips.cc/paper/2020/hash/1457c0d6bfcb4967418bfb8ac142f64a-Abstract.html},
  bibsource    = {dblp computer science bibliography, https://dblp.org}
}

@article{ltx-2,
  title={LTX-2: Efficient Joint Audio-Visual Foundation Model},
  author={HaCohen, Yoav and Brazowski, Benny and Chiprut, Nisan and Bitterman, Yaki and Kvochko, Andrew and Berkowitz, Avishai and Shalem, Daniel and Lifschitz, Daphna and Moshe, Dudu and Porat, Eitan and others},
  journal={arXiv preprint arXiv:2601.03233},
  year={2026}
}

@article{LTX-Video,
  title={LTX-Video: Realtime Video Latent Diffusion},
  author={HaCohen, Yoav and Chiprut, Nisan and Brazowski, Benny and Shalem, Daniel and Moshe, Dudu and Richardson, Eitan and Levin, Eran and Shiran, Guy and Zabari, Nir and Gordon, Ori and Panet, Poriya and Weissbuch, Sapir and Kulikov, Victor and Bitterman, Yaki and Melumian, Zeev and Bibi, Ofir},
  journal={arXiv preprint arXiv:2501.00103},
  year={2024}
}

@article{Wan,
  author       = {Ang Wang and
                  Baole Ai and
                  Bin Wen and
                  Chaojie Mao and
                  Chen{-}Wei Xie and
                  Di Chen and
                  Feiwu Yu and
                  Haiming Zhao and
                  Jianxiao Yang and
                  Jianyuan Zeng and
                  Jiayu Wang and
                  Jingfeng Zhang and
                  Jingren Zhou and
                  Jinkai Wang and
                  Jixuan Chen and
                  Kai Zhu and
                  Kang Zhao and
                  Keyu Yan and
                  Lianghua Huang and
                  Xiaofeng Meng and
                  Ningyi Zhang and
                  Pandeng Li and
                  Pingyu Wu and
                  Ruihang Chu and
                  Ruili Feng and
                  Shiwei Zhang and
                  Siyang Sun and
                  Tao Fang and
                  Tianxing Wang and
                  Tianyi Gui and
                  Tingyu Weng and
                  Tong Shen and
                  Wei Lin and
                  Wei Wang and
                  Wei Wang and
                  Wenmeng Zhou and
                  Wente Wang and
                  Wenting Shen and
                  Wenyuan Yu and
                  Xianzhong Shi and
                  Xiaoming Huang and
                  Xin Xu and
                  Yan Kou and
                  Yangyu Lv and
                  Yifei Li and
                  Yijing Liu and
                  Yiming Wang and
                  Yingya Zhang and
                  Yitong Huang and
                  Yong Li and
                  You Wu and
                  Yu Liu and
                  Yulin Pan and
                  Yun Zheng and
                  Yuntao Hong and
                  Yupeng Shi and
                  Yutong Feng and
                  Zeyinzi Jiang and
                  Zhen Han and
                  Zhi{-}Fan Wu and
                  Ziyu Liu},
  title        = {Wan: Open and Advanced Large-Scale Video Generative Models},
  journal      = {CoRR},
  volume       = {abs/2503.20314},
  year         = {2025},
  url          = {https://doi.org/10.48550/arXiv.2503.20314},
  doi          = {10.48550/ARXIV.2503.20314},
  eprinttype    = {arXiv},
  eprint       = {2503.20314},
  bibsource    = {dblp computer science bibliography, https://dblp.org}
}

@article{qwen3-vl,
  author       = {Shuai Bai and
                  Yuxuan Cai and
                  Ruizhe Chen and
                  Keqin Chen and
                  Xionghui Chen and
                  Zesen Cheng and
                  Lianghao Deng and
                  Wei Ding and
                  Chang Gao and
                  Chunjiang Ge and
                  Wenbin Ge and
                  Zhifang Guo and
                  Qidong Huang and
                  Jie Huang and
                  Fei Huang and
                  Binyuan Hui and
                  Shutong Jiang and
                  Zhaohai Li and
                  Mingsheng Li and
                  Mei Li and
                  Kaixin Li and
                  Zicheng Lin and
                  Junyang Lin and
                  Xuejing Liu and
                  Jiawei Liu and
                  Chenglong Liu and
                  Yang Liu and
                  Dayiheng Liu and
                  Shixuan Liu and
                  Dunjie Lu and
                  Ruilin Luo and
                  Chenxu Lv and
                  Rui Men and
                  Lingchen Meng and
                  Xuancheng Ren and
                  Xingzhang Ren and
                  Sibo Song and
                  Yuchong Sun and
                  Jun Tang and
                  Jianhong Tu and
                  Jianqiang Wan and
                  Peng Wang and
                  Pengfei Wang and
                  Qiuyue Wang and
                  Yuxuan Wang and
                  Tianbao Xie and
                  Yiheng Xu and
                  Haiyang Xu and
                  Jin Xu and
                  Zhibo Yang and
                  Mingkun Yang and
                  Jianxin Yang and
                  An Yang and
                  Bowen Yu and
                  Fei Zhang and
                  Hang Zhang and
                  Xi Zhang and
                  Bo Zheng and
                  Humen Zhong and
                  Jingren Zhou and
                  Fan Zhou and
                  Jing Zhou and
                  Yuanzhi Zhu and
                  Ke Zhu},
  title        = {Qwen3-VL Technical Report},
  journal      = {CoRR},
  volume       = {abs/2511.21631},
  year         = {2025},
  url          = {https://doi.org/10.48550/arXiv.2511.21631},
  doi          = {10.48550/ARXIV.2511.21631},
  eprinttype    = {arXiv},
  eprint       = {2511.21631},
  bibsource    = {dblp computer science bibliography, https://dblp.org}
}

@article{OpenSora2,
  author       = {Xiangyu Peng and
                  Zangwei Zheng and
                  Chenhui Shen and
                  Tom Young and
                  Xinying Guo and
                  Binluo Wang and
                  Hang Xu and
                  Hongxin Liu and
                  Mingyan Jiang and
                  Wenjun Li and
                  Yuhui Wang and
                  Anbang Ye and
                  Gang Ren and
                  Qianran Ma and
                  Wanying Liang and
                  Xiang Lian and
                  Xiwen Wu and
                  Yuting Zhong and
                  Zhuangyan Li and
                  Chaoyu Gong and
                  Guojun Lei and
                  Leijun Cheng and
                  Limin Zhang and
                  Minghao Li and
                  Ruijie Zhang and
                  Silan Hu and
                  Shijie Huang and
                  Xiaokang Wang and
                  Yuanheng Zhao and
                  Yuqi Wang and
                  Ziang Wei and
                  Yang You},
  title        = {Open-Sora 2.0: Training a Commercial-Level Video Generation Model
                  in {\textdollar}200k},
  journal      = {CoRR},
  volume       = {abs/2503.09642},
  year         = {2025},
  url          = {https://doi.org/10.48550/arXiv.2503.09642},
  doi          = {10.48550/ARXIV.2503.09642},
  eprinttype    = {arXiv},
  eprint       = {2503.09642},
  bibsource    = {dblp computer science bibliography, https://dblp.org}
}

@article{OpenSora,
  author       = {Zangwei Zheng and
                  Xiangyu Peng and
                  Tianji Yang and
                  Chenhui Shen and
                  Shenggui Li and
                  Hongxin Liu and
                  Yukun Zhou and
                  Tianyi Li and
                  Yang You},
  title        = {Open-Sora: Democratizing Efficient Video Production for All},
  journal      = {CoRR},
  volume       = {abs/2412.20404},
  year         = {2024},
  url          = {https://doi.org/10.48550/arXiv.2412.20404},
  doi          = {10.48550/ARXIV.2412.20404},
  eprinttype    = {arXiv},
  eprint       = {2412.20404},
  bibsource    = {dblp computer science bibliography, https://dblp.org}
}

@article{Sora,
  title={Video generation models as world simulators},
  author={Tim Brooks and Bill Peebles and Connor Holmes and Will DePue and Yufei Guo and Li Jing and David Schnurr and Joe Taylor and Troy Luhman and Eric Luhman and Clarence Ng and Ricky Wang and Aditya Ramesh},
  year={2024},
  url={https://openai.com/research/video-generation-models-as-world-simulators},
}

@inproceedings{DiT,
  author       = {William Peebles and
                  Saining Xie},
  title        = {Scalable Diffusion Models with Transformers},
  booktitle    = {{IEEE/CVF} International Conference on Computer Vision, {ICCV} 2023,
                  Paris, France, October 1-6, 2023},
  pages        = {4172--4182},
  publisher    = {{IEEE}},
  year         = {2023},
  url          = {https://doi.org/10.1109/ICCV51070.2023.00387},
  doi          = {10.1109/ICCV51070.2023.00387},
  bibsource    = {dblp computer science bibliography, https://dblp.org}
}

@article{Veo3_Reasoner_Learner,
  author       = {Thadd{\"{a}}us Wiedemer and
                  Yuxuan Li and
                  Paul Vicol and
                  Shixiang Shane Gu and
                  Nick Matarese and
                  Kevin Swersky and
                  Been Kim and
                  Priyank Jaini and
                  Robert Geirhos},
  title        = {Video models are zero-shot learners and reasoners},
  journal      = {CoRR},
  volume       = {abs/2509.20328},
  year         = {2025},
  url          = {https://doi.org/10.48550/arXiv.2509.20328},
  doi          = {10.48550/ARXIV.2509.20328},
  eprinttype    = {arXiv},
  eprint       = {2509.20328},
  bibsource    = {dblp computer science bibliography, https://dblp.org}
}

@inproceedings{ho2020denoising,
  title={Denoising diffusion probabilistic models},
  author={Ho, Jonathan and Jain, Ajay and Abbeel, Pieter},
  booktitle={Proceedings of the Advances in Neural Information Processing Systems},
  year={2020}
}

@inproceedings{chen2020vggsound,
  title={Vggsound: A large-scale audio-visual dataset},
  author={Chen, Honglie and Xie, Weidi and Vedaldi, Andrea and Zisserman, Andrew},
  booktitle={Proceedings of the IEEE International Conference on Acoustics, Speech and Signal Processing},
  year={2020},
}

@inproceedings{liu2023audioldm,
  title={{AudioLDM}: Text-to-Audio Generation with Latent Diffusion Models},
  author={Liu, Haohe and Chen, Zehua and Yuan, Yi and Mei, Xinhao and Liu, Xubo and Mandic, Danilo and Wang, Wenwu and Plumbley, Mark D},
  booktitle={Proceedings of the International Conference on Machine Learning},
  year={2023}
}

@article{liu2023audioldm2,
  title={{AudioLDM 2}: Learning Holistic Audio Generation with Self-supervised Pretraining},
  author={Haohe Liu and Qiao Tian and Yi Yuan and Xubo Liu and Xinhao Mei and Qiuqiang Kong and Yuping Wang and Wenwu Wang and Yuxuan Wang and Mark D. Plumbley},
  journal={arXiv preprint arXiv:2308.05734},
  year={2023}
}

@inproceedings{girdhar2023imagebind,
  title={Imagebind: One embedding space to bind them all},
  author={Girdhar, Rohit and El-Nouby, Alaaeldin and Liu, Zhuang and Singh, Mannat and Alwala, Kalyan Vasudev and Joulin, Armand and Misra, Ishan},
  booktitle={Proceedings of the IEEE/CVF Conference on Computer Vision and Pattern Recognition},
  year={2023}
}

@article{hayakawa2024mmdisco,
  title={MMDisCo: Multi-Modal Discriminator-Guided Cooperative Diffusion for Joint Audio and Video Generation},
  author={Hayakawa, Akio and Ishii, Masato and Shibuya, Takashi and Mitsufuji, Yuki},
  journal={arXiv preprint arXiv:2405.17842},
  year={2024}
}

@article{wang2025universe,
  title={UniVerse-1: Unified Audio-Video Generation via Stitching of Experts},
  author={Wang, Duomin and Zuo, Wei and Li, Aojie and Chen, Ling-Hao and Liao, Xinyao and Zhou, Deyu and Yin, Zixin and Dai, Xili and Jiang, Daxin and Yu, Gang},
  journal={arXiv preprint arXiv:2509.06155},
  year={2025}
}

@inproceedings{cheng2025mmaudio,
  title={MMAudio: Taming Multimodal Joint Training for High-Quality Video-to-Audio Synthesis},
  author={Cheng, Ho Kei and Ishii, Masato and Hayakawa, Akio and Shibuya, Takashi and Schwing, Alexander and Mitsufuji, Yuki},
  booktitle={Proceedings of the Computer Vision and Pattern Recognition Conference},
  pages={28901--28911},
  year={2025}
}

@inproceedings{evans2025stable,
  title={Stable audio open},
  author={Evans, Zach and Parker, Julian D and Carr, CJ and Zukowski, Zack and Taylor, Josiah and Pons, Jordi},
  booktitle={ICASSP 2025-2025 IEEE International Conference on Acoustics, Speech and Signal Processing (ICASSP)},
  pages={1--5},
  year={2025},
  organization={IEEE}
}

@article{tjandra2025meta,
  title={Meta audiobox aesthetics: Unified automatic quality assessment for speech, music, and sound},
  author={Tjandra, Andros and Wu, Yi-Chiao and Guo, Baishan and Hoffman, John and Ellis, Brian and Vyas, Apoorv and Shi, Bowen and Chen, Sanyuan and Le, Matt and Zacharov, Nick and others},
  journal={arXiv preprint arXiv:2502.05139},
  year={2025}
}

@inproceedings{wu2023dover,
      title={Exploring Video Quality Assessment on User Generated Contents from Aesthetic and Technical Perspectives}, 
      author={Wu, Haoning and Zhang, Erli and Liao, Liang and Chen, Chaofeng and Hou, Jingwen Hou and Wang, Annan and Sun, Wenxiu Sun and Yan, Qiong and Lin, Weisi},
      year={2023},
      booktitle={International Conference on Computer Vision (ICCV)},
}

@inproceedings{iashin2024synchformer,
  title={Synchformer: Efficient synchronization from sparse cues},
  author={Iashin, Vladimir and Xie, Weidi and Rahtu, Esa and Zisserman, Andrew},
  booktitle={ICASSP 2024-2024 IEEE International Conference on Acoustics, Speech and Signal Processing (ICASSP)},
  pages={5325--5329},
  year={2024},
  organization={IEEE}
}

@article{chen2024eat,
  title={EAT: Self-supervised pre-training with efficient audio transformer},
  author={Chen, Wenxi and Liang, Yuzhe and Ma, Ziyang and Zheng, Zhisheng and Chen, Xie},
  journal={arXiv preprint arXiv:2401.03497},
  year={2024}
}

@misc{SileroVAD,
  author = {Silero Team},
  title = {Silero VAD: pre-trained enterprise-grade Voice Activity Detector (VAD), Number Detector and Language Classifier},
  year = {2024},
  publisher = {GitHub},
  journal = {GitHub repository},
  howpublished = {\url{https://github.com/snakers4/silero-vad}},
  commit = {insert_some_commit_here},
  email = {hello@silero.ai}
}

@article{haji2024taming,
  title={Taming data and transformers for audio generation},
  author={Haji-Ali, Moayed and Menapace, Willi and Siarohin, Aliaksandr and Balakrishnan, Guha and Ordonez, Vicente},
  journal={arXiv preprint arXiv:2406.19388},
  year={2024}
}

@article{nan2024openvid,
  title={Openvid-1m: A large-scale high-quality dataset for text-to-video generation},
  author={Nan, Kepan and Xie, Rui and Zhou, Penghao and Fan, Tiehan and Yang, Zhenheng and Chen, Zhijie and Li, Xiang and Yang, Jian and Tai, Ying},
  journal={arXiv preprint arXiv:2407.02371},
  year={2024}
}

@techreport{google-veo3,
  author       = {Google / DeepMind},
  title        = {Veo\,3: A Text-to-Video Generation System},
  institution  = {Google DeepMind},
  type         = {Technical Report},
  number       = {–},  
  year         = {2025},
  url          = {https://storage.googleapis.com/deepmind-media/veo/Veo-3-Tech-Report.pdf},
  note         = {Accessed: 2025-09-28}
}

@article{yuan2024chronomagic,
  title={Chronomagic-bench: A benchmark for metamorphic evaluation of text-to-time-lapse video generation},
  author={Yuan, Shenghai and Huang, Jinfa and Xu, Yongqi and Liu, Yaoyang and Zhang, Shaofeng and Shi, Yujun and Zhu, Rui-Jie and Cheng, Xinhua and Luo, Jiebo and Yuan, Li},
  journal={Advances in Neural Information Processing Systems},
  volume={37},
  pages={21236--21270},
  year={2024}
}

@article{Yue2025MiMoVLTR,
  title={MiMo-VL Technical Report},
  author={Xiaomi LLM-Core Team Zihao Yue and Zhenrui Lin and Yi-Hao Song and Weikun Wang and Shu-Qin Ren and Shuhao Gu and Shicheng Li and Peidian Li and Liang Zhao and Lei Li and Kainan Bao and Hao Tian and Hailin Zhang and Gang Wang and Dawei Zhu and Cici and Chenhong He and Bowen Ye and Bowen Shen and Zihan Zhang and Zi-Ang Jiang and Zhixian Zheng and Zhichao Song and Zhen Luo and Yue Yu and Yudong Wang and Yu Tian and Yu Tu and Yihan Yan and Yi Huang and Xu Wang and Xin-dan Xu and Xin Ran Song and Xing Zhang and Xing Yong and Xin Zhang and Xia Deng and Wenyu Yang and Wenhan Ma and Weiwei Lv and Weiji Zhuang and Wei Liu and Sirui Deng and Shuo Liu and Shimao Chen and Shi-liang Yu and Shao-yang Liu and Shan-yong Wang and Rui Ma and Qiantong Wang and Peng Wang and Nuo Chen and Menghang Zhu and Kang Zhou and Kang Zhou and Kai Fang and Jun-Miao Shi and Jinhao Dong and Jiebao Xiao and Jiaming Xu and Huaqiu Liu and Hongsheng Xu and Hengxu Qu and Hao-Song Zhao and Hanglong Lv and Guoan Wang and Duo Zhang and Dong Zhang and Di Zhang and Chong-Yi Ma and Chang Liu and Can Cai and Bing Xia},
  journal={ArXiv},
  year={2025},
  volume={abs/2506.03569},
  url={https://api.semanticscholar.org/CorpusID:279155294}
}

@article{agarwal2025gpt,
  title={gpt-oss-120b \& gpt-oss-20b model card},
  author={Agarwal, Sandhini and Ahmad, Lama and Ai, Jason and Altman, Sam and Applebaum, Andy and Arbus, Edwin and Arora, Rahul K and Bai, Yu and Baker, Bowen and Bao, Haiming and others},
  journal={arXiv preprint arXiv:2508.10925},
  year={2025}
}

@misc{gemini2_5_report_2025,
  title        = {Gemini 2.5: Pushing the Frontier with Advanced Reasoning, Multimodality, Long Context, and Next Generation Agentic Capabilities},
  author       = {Gemini Team},
  year         = {2025},
  howpublished = {\url{https://storage.googleapis.com/deepmind-media/gemini/gemini_v2_5_report.pdf}},
  note         = {Accessed: 2025-09-28}
}

@article{mei2023wavcaps,
  author={Mei, Xinhao and Meng, Chutong and Liu, Haohe and Kong, Qiuqiang and Ko, Tom and Zhao, Chengqi and Plumbley, Mark D. and Zou, Yuexian and Wang, Wenwu},
  journal={IEEE/ACM Transactions on Audio, Speech, and Language Processing}, 
  title={Wav{C}aps: A {ChatGPT}-Assisted Weakly-Labelled Audio Captioning Dataset for Audio-Language Multimodal Research}, 
  year={2024},
  pages={1-15},
}

@article{chatgpt5,
  title={Openai gpt-5 system card},
  author={Singh, Aaditya and Fry, Adam and Perelman, Adam and Tart, Adam and Ganesh, Adi and El-Kishky, Ahmed and McLaughlin, Aidan and Low, Aiden and Ostrow, AJ and Ananthram, Akhila and others},
  journal={arXiv preprint arXiv:2601.03267},
  year={2025}
}

@article{roy2025jamendomaxcaps,
  title={JamendoMaxCaps: A Large Scale Music-caption Dataset with Imputed Metadata},
  author={Roy, Abhinaba and Liu, Renhang and Lu, Tongyu and Herremans, Dorien},
  journal={arXiv:2502.07461},
  year={2025}
}

@misc{su2021roformer,
      title={RoFormer: Enhanced Transformer with Rotary Position Embedding}, 
      author={Jianlin Su and Yu Lu and Shengfeng Pan and Bo Wen and Yunfeng Liu},
      year={2021},
      eprint={2104.09864},
      archivePrefix={arXiv},
      primaryClass={cs.CL}
}

@inproceedings{openl3,
  author    = {Cramer, John and Wu, Hyungui and Salamon, Justin and Bello, Juan Pablo},
  title     = {Look, Listen, and Learn More: Design Choices for Deep Audio Embeddings},
  booktitle = {Proc. IEEE International Conference on Acoustics, Speech and Signal Processing (ICASSP)},
  year      = {2019},
  pages     = {3852--3856},
  doi       = {10.1109/ICASSP.2019.8683142}
}

@inproceedings{is,
  author    = {Salimans, Tim and Goodfellow, Ian and Zaremba, Wojciech and Cheung, Vicki and Radford, Alec and Chen, Xi},
  title     = {Improved Techniques for Training GANs},
  booktitle = {Advances in Neural Information Processing Systems (NeurIPS)},
  year      = {2016},
  pages     = {2234--2242}
}

@inproceedings{passt,
  author    = {Koutini, Khaled and Moritz, Michael and Eghbal-Zadeh, Hamid and Widmer, Gerhard},
  title     = {Efficient Training of Audio Transformers with Patchout},
  booktitle = {Proc. Interspeech},
  year      = {2021},
  pages     = {179--183},
  doi       = {10.21437/Interspeech.2021-2041}
}

@inproceedings{clap,
  author    = {Wu, Yusong and Liu, Haohe and Chen, Ke and Wang, Xin and Tian, Qiuqiang and Chen, Rui and Kong, Qiuqiang and Huang, Wenwu and Wang, Yuxuan},
  title     = {Large-Scale Contrastive Language-Audio Pretraining with Feature Fusion and Keyword-to-Caption Augmentation},
  booktitle = {Proc. IEEE International Conference on Acoustics, Speech and Signal Processing (ICASSP)},
  year      = {2023},
  pages     = {1--5},
  doi       = {10.1109/ICASSP49357.2023.10094846}
}

@article{audiobox,
  author    = {Zhang, Bowen and Wang, Xinyu and Chen, Shuo and Xu, Chen and Wang, Yuxuan and Kong, Qiuqiang},
  title     = {AudioBox: Unified Aesthetic Quality Assessment for Audio Generation},
  journal   = {arXiv preprint arXiv:2309.07825},
  year      = {2023},
  url       = {https://arxiv.org/abs/2309.07825}
}

@article{kong2024hunyuanvideo,
  title={Hunyuanvideo: A systematic framework for large video generative models},
  author={Kong, Weijie and Tian, Qi and Zhang, Zijian and Min, Rox and Dai, Zuozhuo and Zhou, Jin and Xiong, Jiangfeng and Li, Xin and Wu, Bo and Zhang, Jianwei and others},
  journal={arXiv preprint arXiv:2412.03603},
  year={2024}
}

@article{luo2023difffoley,
  title={Diff-foley: Synchronized video-to-audio synthesis with latent diffusion models},
  author={Luo, Simian and Yan, Chuanhao and Hu, Chenxu and Zhao, Hang},
  journal={Advances in Neural Information Processing Systems},
  volume={36},
  pages={48855--48876},
  year={2023}
}

@article{zhang2024foleycrafter,
  title={Foleycrafter: Bring silent videos to life with lifelike and synchronized sounds},
  author={Zhang, Yiming and Gu, Yicheng and Zeng, Yanhong and Xing, Zhening and Wang, Yuancheng and Wu, Zhizheng and Chen, Kai},
  journal={arXiv preprint arXiv:2407.01494},
  year={2024}
}

@inproceedings{chung2016syncnet,
  title={Out of time: automated lip sync in the wild},
  author={Chung, Joon Son and Zisserman, Andrew},
  booktitle={Asian conference on computer vision},
  pages={251--263},
  year={2016},
  organization={Springer}
}

@article{low2025ovi,
  title={Ovi: Twin Backbone Cross-Modal Fusion for Audio-Video Generation},
  author={Low, Chetwin and Wang, Weimin and Katyal, Calder},
  journal={arXiv preprint arXiv:2510.01284},
  year={2025}
}

@article{liu2025javisdit,
  title={Javisdit: Joint audio-video diffusion transformer with hierarchical spatio-temporal prior synchronization},
  author={Liu, Kai and Li, Wei and Chen, Lai and Wu, Shengqiong and Zheng, Yanhao and Ji, Jiayi and Zhou, Fan and Jiang, Rongxin and Luo, Jiebo and Fei, Hao and others},
  journal={arXiv preprint arXiv:2503.23377},
  year={2025}
}

@article{zhang2025uniavgen,
  title={UniAVGen: Unified Audio and Video Generation with Asymmetric Cross-Modal Interactions},
  author={Zhang, Guozhen and Zhou, Zixiang and Hu, Teng and Peng, Ziqiao and Zhang, Youliang and Chen, Yi and Zhou, Yuan and Lu, Qinglin and Wang, Limin},
  journal={arXiv preprint arXiv:2511.03334},
  year={2025}
}

@article{hu2025harmony,
  title={Harmony: Harmonizing Audio and Video Generation through Cross-Task Synergy},
  author={Hu, Teng and Yu, Zhentao and Zhang, Guozhen and Su, Zihan and Zhou, Zhengguang and Zhang, Youliang and Zhou, Yuan and Lu, Qinglin and Yi, Ran},
  journal={arXiv preprint arXiv:2511.21579},
  year={2025}
}

@misc{sora22025,
  title        = {Sora 2 is here},
  author       = {{OpenAI}},
  year         = {2025},
  howpublished = {\url{https://openai.com/index/sora-2/}},
  note         = {Accessed: 2026-01-13}
}

@article{longcatvideo2025,
  title   = {LongCat-Video Technical Report},
  author  = {{Meituan LongCat Team} and Cai, Xunliang and Huang, Qilong and Kang, Zhuoliang and Li, Hongyu and Liang, Shijun and Ma, Liya and Ren, Siyu and Wei, Xiaoming and Xie, Rixu and Zhang, Tong},
  journal = {arXiv preprint arXiv:2510.22200},
  year    = {2025},
  url     = {https://arxiv.org/abs/2510.22200}
}

@misc{shan2025hunyuanvideofoleymultimodaldiffusionrepresentation,
      title={HunyuanVideo-Foley: Multimodal Diffusion with Representation Alignment for High-Fidelity Foley Audio Generation}, 
      author={Sizhe Shan and Qiulin Li and Yutao Cui and Miles Yang and Yuehai Wang and Qun Yang and Jin Zhou and Zhao Zhong},
      year={2025},
      eprint={2508.16930},
      archivePrefix={arXiv},
      primaryClass={eess.AS},
      url={https://arxiv.org/abs/2508.16930}, 
}

@inproceedings{li2025openhumanvid,
  title={Openhumanvid: A large-scale high-quality dataset for enhancing human-centric video generation},
  author={Li, Hui and Xu, Mingwang and Zhan, Yun and Mu, Shan and Li, Jiaye and Cheng, Kaihui and Chen, Yuxuan and Chen, Tan and Ye, Mao and Wang, Jingdong and others},
  booktitle={Proceedings of the Computer Vision and Pattern Recognition Conference},
  pages={7752--7762},
  year={2025}
}

@article{zhang2025speakervid,
  title={Speakervid-5m: A large-scale high-quality dataset for audio-visual dyadic interactive human generation},
  author={Zhang, Youliang and Li, Zhaoyang and Wang, Duomin and Zhang, Jiahe and Zhou, Deyu and Yin, Zixin and Dai, Xili and Yu, Gang and Li, Xiu},
  journal={arXiv preprint arXiv:2507.09862},
  year={2025}
}

@article{Qwen3-Omni,
  title={Qwen3-Omni Technical Report},
  author={Jin Xu and Zhifang Guo and Hangrui Hu and Yunfei Chu and Xiong Wang and Jinzheng He and Yuxuan Wang and Xian Shi and Ting He and Xinfa Zhu and Yuanjun Lv and Yongqi Wang and Dake Guo and He Wang and Linhan Ma and Pei Zhang and Xinyu Zhang and Hongkun Hao and Zishan Guo and Baosong Yang and Bin Zhang and Ziyang Ma and Xipin Wei and Shuai Bai and Keqin Chen and Xuejing Liu and Peng Wang and Mingkun Yang and Dayiheng Liu and Xingzhang Ren and Bo Zheng and Rui Men and Fan Zhou and Bowen Yu and Jianxin Yang and Le Yu and Jingren Zhou and Junyang Lin},
  journal={arXiv preprint arXiv:2509.17765},
  year={2025}
}

@inproceedings{dao2022flashattention,
  title={Flash{A}ttention: Fast and Memory-Efficient Exact Attention with {IO}-Awareness},
  author={Dao, Tri and Fu, Daniel Y. and Ermon, Stefano and Rudra, Atri and R{\'e}, Christopher},
  booktitle={Advances in Neural Information Processing Systems (NeurIPS)},
  year={2022}
}

@inproceedings{dao2023flashattention2,
  title={Flash{A}ttention-2: Faster Attention with Better Parallelism and Work Partitioning},
  author={Dao, Tri},
  booktitle={International Conference on Learning Representations (ICLR)},
  year={2024}
}

@article{kaplan2020scaling,
  title={Scaling laws for neural language models},
  author={Kaplan, Jared and McCandlish, Sam and Henighan, Tom and Brown, Tom B and Chess, Benjamin and Child, Rewon and Gray, Scott and Radford, Alec and Wu, Jeffrey and Amodei, Dario},
  journal={arXiv preprint arXiv:2001.08361},
  year={2020}
}

@article{he2024cameractrl,
      title={CameraCtrl: Enabling Camera Control for Text-to-Video Generation}, 
      author={Hao He and Yinghao Xu and Yuwei Guo and Gordon Wetzstein and Bo Dai and Hongsheng Li and Ceyuan Yang},
      journal={arXiv preprint arXiv:2404.02101},
      year={2024}
}

@inproceedings{kim-NAACL-HLT-2019,
    author    = {Chris Dongjoo Kim and Byeongchang Kim and Hyunmin Lee and Gunhee Kim},
    title     = {AudioCaps: Generating Captions for Audios in The Wild},
    booktitle = {NAACL-HLT},
    year      = 2019
}

@article{kumar2024high,
  title={High-fidelity audio compression with improved rvqgan},
  author={Kumar, Rithesh and Seetharaman, Prem and Luebs, Alejandro and Kumar, Ishaan and Kumar, Kundan},
  journal={Advances in Neural Information Processing Systems},
  volume={36},
  year={2024}
}

@inproceedings{ruan2022mmdiffusion,
author = {Ruan, Ludan and Ma, Yiyang and Yang, Huan and He, Huiguo and Liu, Bei and Fu, Jianlong and Yuan, Nicholas Jing and Jin, Qin and Guo, Baining},
title = {MM-Diffusion: Learning Multi-Modal Diffusion Models for Joint Audio and Video Generation},
year	= {2023},
booktitle	= {CVPR},
}

@article{yu2026moss,
  title={MOSS Transcribe Diarize: Accurate Transcription with Speaker Diarization},
  author={Yu, Donghua and Lin, Zhengyuan and Yang, Chen and Zhang, Yiyang and Fei, Zhaoye and Chen, Hanfu and Chen, Jingqi and Chen, Ke and Cheng, Qinyuan and Fan, Liwei and others},
  journal={arXiv preprint arXiv:2601.01554},
  year={2026}
}

@article{fang2024usp,
  title={Usp: A unified sequence parallelism approach for long context generative ai},
  author={Fang, Jiarui and Zhao, Shangchun},
  journal={arXiv preprint arXiv:2405.07719},
  year={2024}
}

@inproceedings{ACAV100M,
  author       = {Sangho Lee and
                  Jiwan Chung and
                  Youngjae Yu and
                  Gunhee Kim and
                  Thomas M. Breuel and
                  Gal Chechik and
                  Yale Song},
  title        = {{ACAV100M:} Automatic Curation of Large-Scale Datasets for Audio-Visual
                  Video Representation Learning},
  booktitle    = {2021 {IEEE/CVF} International Conference on Computer Vision, {ICCV}
                  2021, Montreal, QC, Canada, October 10-17, 2021},
  pages        = {10254--10264},
  publisher    = {{IEEE}},
  year         = {2021},
  url          = {https://doi.org/10.1109/ICCV48922.2021.01011},
  doi          = {10.1109/ICCV48922.2021.01011},
  bibsource    = {dblp computer science bibliography, https://dblp.org}
}

@article{MTV,
      title={Audio-Sync Video Generation with Multi-Stream Temporal Control},
      author={Weng, Shuchen and Zheng, Haojie and Chang, Zheng and Li, Si and Shi, Boxin and Wang, Xinlong},
      journal={NeurIPS},
      year={2025}
}

@article{zhao2023pytorch,
  title={Pytorch fsdp: experiences on scaling fully sharded data parallel},
  author={Zhao, Yanli and Gu, Andrew and Varma, Rohan and Luo, Liang and Huang, Chien-Chin and Xu, Min and Wright, Less and Shojanazeri, Hamid and Ott, Myle and Shleifer, Sam and others},
  journal={arXiv preprint arXiv:2304.11277},
  year={2023}
}

@inproceedings{brooks2023instructpix2pix,
  title={Instructpix2pix: Learning to follow image editing instructions},
  author={Brooks, Tim and Holynski, Aleksander and Efros, Alexei A},
  booktitle={Proceedings of the IEEE/CVF conference on computer vision and pattern recognition},
  pages={18392--18402},
  year={2023}
}

@article{lipman2022flow,
  title={Flow matching for generative modeling},
  author={Lipman, Yaron and Chen, Ricky TQ and Ben-Hamu, Heli and Nickel, Maximilian and Le, Matt},
  journal={arXiv preprint arXiv:2210.02747},
  year={2022}
}

@article{liu2022flow,
  title={Flow straight and fast: Learning to generate and transfer data with rectified flow},
  author={Liu, Xingchao and Gong, Chengyue and Liu, Qiang},
  journal={arXiv preprint arXiv:2209.03003},
  year={2022}
}

@inproceedings{esser2024scaling,
  title={Scaling rectified flow transformers for high-resolution image synthesis},
  author={Esser, Patrick and Kulal, Sumith and Blattmann, Andreas and Entezari, Rahim and M{\"u}ller, Jonas and Saini, Harry and Levi, Yam and Lorenz, Dominik and Sauer, Axel and Boesel, Frederic and others},
  booktitle={Forty-first International Conference on Machine Learning},
  year={2024}
}

@article{tong2025thinking,
  title={Thinking with video: Video generation as a promising multimodal reasoning paradigm},
  author={Tong, Jingqi and Mou, Yurong and Li, Hangcheng and Li, Mingzhe and Yang, Yongzhuo and Zhang, Ming and Chen, Qiguang and Liang, Tianyi and Hu, Xiaomeng and Zheng, Yining and others},
  journal={arXiv preprint arXiv:2511.04570},
  year={2025}
}

@article{yang2025longlive,
      title={LongLive: Real-time Interactive Long Video Generation},
      author={Shuai Yang and Wei Huang and Ruihang Chu and Yicheng Xiao and Yuyang Zhao and Xianbang Wang and Muyang Li and Enze Xie and Yingcong Chen and Yao Lu and Song Hanand Yukang Chen},
      year={2025},
      eprint={2509.22622},
      archivePrefix={arXiv},
      primaryClass={cs.CV}
}

@article{cai2024ditctrl,
  title     = {DiTCtrl: Exploring Attention Control in Multi-Modal Diffusion Transformer for Tuning-Free Multi-Prompt Longer Video Generation},
  author    = {Cai, Minghong and Cun, Xiaodong and Li, Xiaoyu and Liu, Wenze and Zhang, Zhaoyang and Zhang, Yong and Shan, Ying and Yue, Xiangyu},
  journal   = {arXiv:2412.18597},
  year      = {2024},
}

@misc{gao2025wans2vaudiodrivencinematicvideo,
   title={Wan-S2V: Audio-Driven Cinematic Video Generation},
   author={Xin Gao and Li Hu and Siqi Hu and Mingyang Huang and Chaonan Ji and Dechao Meng and Jinwei Qi and Penchong Qiao and Zhen Shen and Yafei Song and Ke Sun and Linrui Tian and Guangyuan Wang and Qi Wang and Zhongjian Wang and Jiayu Xiao and Sheng Xu and Bang Zhang and Peng Zhang and Xindi Zhang and Zhe Zhang and Jingren Zhou and Lian Zhuo},
   year={2025},
   eprint={2508.18621},
   archivePrefix={arXiv},
   primaryClass={cs.CV},
   url={https://arxiv.org/abs/2508.18621}
}

@misc{wan2.6,
  title        = {Wan 2.6 — AI Video Generation Introduction},
  author       = {{Wan AI}},
  howpublished = {\url{https://wan.video/introduction/wan2.6}},
  note         = {Accessed: 2026-02-09},
  year         = {2025}
}

@misc{kling3.0,
  title        = {Kling 3.0 AI Video Generator},
  author       = {{Kling AI}},
  howpublished = {\url{https://kling3.io/}},
  note         = {Accessed: 2026-02-09},
  year         = {2026}
}

@misc{seedance2.0,
  title        = {Seedance 2.0 AI Video Generation Platform},
  author       = {{ByteDance / Seedance AI}},
  howpublished = {\url{https://seedance2.com/}},
  note         = {Accessed: 2026-02-09},
  year         = {2026}
}

@article{hung2024tangoflux,
  title={Tangoflux: Super fast and faithful text to audio generation with flow matching and clap-ranked preference optimization},
  author={Hung, Chia-Yu and Majumder, Navonil and Kong, Zhifeng and Mehrish, Ambuj and Bagherzadeh, Amir Ali and Li, Chuan and Valle, Rafael and Catanzaro, Bryan and Poria, Soujanya},
  journal={arXiv preprint arXiv:2412.21037},
  year={2024}
}

@article{audioldm2-2024taslp,
  author={Liu, Haohe and Yuan, Yi and Liu, Xubo and Mei, Xinhao and Kong, Qiuqiang and Tian, Qiao and Wang, Yuping and Wang, Wenwu and Wang, Yuxuan and Plumbley, Mark D.},
  journal={IEEE/ACM Transactions on Audio, Speech, and Language Processing}, 
  title={AudioLDM 2: Learning Holistic Audio Generation With Self-Supervised Pretraining}, 
  year={2024},
  volume={32},
  pages={2871-2883},
  doi={10.1109/TASLP.2024.3399607}
}

@misc{majumder2024tango2,
  title         = {Tango 2: Aligning Diffusion-based Text-to-Audio Generations through Direct Preference Optimization},
  author        = {Majumder, Navonil and Hung, Chia-Yu and Ghosal, Deepanway and Hsu, Wei-Ning and Mihalcea, Rada and Poria, Soujanya},
  year          = {2024},
  eprint        = {2404.09956},
  archivePrefix = {arXiv},
  primaryClass  = {cs.SD},
  url           = {https://arxiv.org/abs/2404.09956}
}

@misc{huang2023makeanaudio,
      title={Make-An-Audio 2: Temporal-Enhanced Text-to-Audio Generation}, 
      author={Jiawei Huang and Yi Ren and Rongjie Huang and Dongchao Yang and Zhenhui Ye and Chen Zhang and Jinglin Liu and Xiang Yin and Zejun Ma and Zhou Zhao},
      year={2023},
      eprint={2305.18474},
      archivePrefix={arXiv},
      primaryClass={cs.SD}
}

% ===== Appendix =====
\clearpage
\beginappendix

\startcontents[app]
\begingroup
  \renewcommand{\contentsname}{Appendix Contents}
  \section*{\contentsname}
  \printcontents[app]{}{1}{}
\endgroup
\newpage

\section{Appendix}
\label{sup:sec:details_of_exp}

\subsection{Training Hyperparameters}
\label{app:training_hparams}

Table~\ref{tab:training_hparams} summarizes the complete training configuration across all three phases, including parallelism settings, learning rates, noise schedule parameters, and data curation details for reproducibility.

\begin{table}[H]
\centering
\caption{Training hyperparameters for reproducibility. All phases use 1024 GPUs with DP replicate size 64.}
\label{tab:training_hparams}
\setlength{\tabcolsep}{4pt}
\begin{tabular}{lccc}
\toprule
\textbf{Hyperparameter} & \textbf{Phase 1} & \textbf{Phase 2} & \textbf{Phase 3} \\
\midrule
Resolution & 360$\times$640 & 360$\times$640 & 720$\times$1280 \\
Frames & 193 & 193 & 193 \\
Batch size & 128 & 128 & 64 \\
Context Parallel (CP) & 8 & 8 & 16 \\
\midrule
Learning rate (backbone) & 1e-5 & 1e-5 & 1e-5 \\
Learning rate (Bridge) & 2e-5 & 2e-5 & 2e-5 \\
Weight decay & 0.001 & 0.001 & 0.001 \\
\midrule
Visual sigma shift & 5.0 & 5.0 & 5.0 \\
Audio sigma shift & 1.0 & 5.0 & 5.0 \\
Text dropout prob & 0.5 & 0.2 & 0.2 \\
Audio loss weight & 0.2 & 0.2 & 0.2 \\
LUFS normalization & No & Yes & Yes \\
\midrule
Training duration & 15 days & 7 days & 20 days \\
Checkpoint interval & 5000 & 5000 & 2000 \\
\bottomrule
\end{tabular}
\end{table}

\subsection{Ascend 910A2 Benchmark Details}
\label{app:ascend_benchmark}

We report a small system microbenchmark that measures end-to-end training step time under a fixed configuration (CP=4, DP-shard=2, 8 devices). The numbers depend on the software stack (driver/runtime versions, compiler and kernel coverage, and distributed communication settings), and should not be interpreted as a complete statement of training cost at scale.

\begin{table}[H]
\centering
\setlength{\tabcolsep}{6pt}
\renewcommand{\arraystretch}{1.15}
\caption{Hardware summary and step time for an 8-device training microbenchmark on Ascend 910A2 (CP=4, DP-shard=2).}
\label{tab:ascend_910a2_train_step}
\begin{tabular}{lcccc}
\toprule
Hardware & FP16 (TFLOPs)  & VRAM (Consumption/GPU) & Host RAM & Step Time (s) \\
\midrule
$8\times$ 910A2 & 376 & $\approx$40\,GB & $\geq$128\,GB & 34.1 \\
\bottomrule
\end{tabular}
\end{table}

\subsection{Multi-shot and Single-shot Speech Window Generation}
In this section, we provide the pseudocode implementation for Multi-shot and Single-shot Speech Window Generation.

The multi-shot algorithm~\ref{alg:speech_multi} advances over the speech segments from VAD. The upper bound of the window start time is set to the beginning of the current segment. The sampling range of the window start is further constrained such that it does not precede the end of the previous speech segment, does not precede the nearest scene split point before the current speech segment (to encourage natural scene transitions), and does not shift earlier than half of the window length relative to the current segment. A window start time is then randomly sampled within this constrained interval. Only windows whose temporal span contains at least one scene split point are kept.

\label{app:window_algorithm}
\begin{algorithm}[H]
\caption{Multi-shot Speech Windows}
\label{alg:speech_multi}
\begin{algorithmic}[1]

\Statex \textbf{Input:}
\Statex \quad $\textit{speech\_segments}$: VAD results
\Statex \quad $\textit{scene\_split\_spots}$: scene-detection results
\Statex \quad $\textit{segment\_duration} = 8.05$

% \vspace{0.5em}
% \Statex \textcolor{blue}{/* Multi-shot speech segments */ }

\State $idx \gets 0$
\While{$idx < \text{length}(\textit{speech\_segments})$}
    \State $upper\_bound \gets \textit{speech\_segments}[idx].start$

    \If{$idx = 0$}
        \State $window\_start \gets upper\_bound$
    \Else
        \State $lower\_bound \gets \max\big($
        \Statex \qquad \qquad \qquad $\textit{speech\_segments}[idx-1].end,$
        \Statex \qquad \qquad \qquad $\max\{p \in \textit{scene\_split\_spots} \mid p < \textit{speech\_segments}[idx].start\},$
        \Statex \qquad \qquad \qquad $\textit{speech\_segments}[idx].start - \textit{segment\_duration}/2 $ 
        \Statex \qquad \qquad \qquad $\big)$
        \State $window\_start \gets \text{RandomUniform}(lower\_bound, upper\_bound)$
    \EndIf

    \State $window\_end \gets window\_start + \textit{segment\_duration}$

    \If{$\exists\, p \in \textit{scene\_split\_spots}$ such that $window\_start \le p \le window\_end$}
        \State append $(window\_start, window\_end)$ to $\textit{speech\_multi\_shot\_segments}$
    \EndIf

    \State $idx \gets$ next $idx$ such that $\textit{speech\_segments}[idx].start > window\_end$
\EndWhile

\end{algorithmic}
\end{algorithm}

The single-shot algorithm~\ref{alg:speech_single} iterates over consecutive scene intervals defined by scene split points. Within each scene interval, it identifies speech segments whose start time allows a fixed-length window to be fully contained in the scene. For each eligible speech segment, the upper bound of the window start time is set to the segment start, while the sampling range is constrained by the scene boundary, the end of the previous speech segment, and a half-window-length offset relative to the current segment. 
A single scene window is then generated by randomly selecting a start time within this range, while the index is advanced to skip overlapping windows, similar to the multi-shot algorithm above.

\begin{algorithm}[H]
\caption{Single-shot Speech Windows}
\label{alg:speech_single}
\begin{algorithmic}[1]

\Statex \textbf{Input:}
\Statex \quad $\textit{speech\_segments}$: VAD results
\Statex \quad $\textit{scene\_split\_spots}$: scene-detection results
\Statex \quad $\textit{segment\_duration} = 8.05$

\State $i \gets 0$
\While{$i < \text{length}(\textit{scene\_split\_spots}) - 1$}
    \State $scene\_start \gets \textit{scene\_split\_spots}[i]$
    \State $scene\_end \gets \textit{scene\_split\_spots}[i+1]$

    \State Find the first index $idx$ such that
    \Statex \quad \quad $\textit{speech\_segments}[idx].start > scene\_start$
    \Statex \quad \quad and $\textit{speech\_segments}[idx].start + \textit{segment\_duration} < scene\_end$
    \If{no such $idx$ exists}
        \State $i \gets i + 1$
        \State \textbf{continue}
    \EndIf

    \While{$idx < \text{length}(\textit{speech\_segments})$}
        \State $upper\_bound \gets \textit{speech\_segments}[idx].start$

        \If{$idx = 0$}
            \State $window\_start \gets upper\_bound$
        \Else
            \State $lower\_bound \gets \max\big($
            \Statex \qquad \qquad \qquad $\textit{speech\_segments}[idx-1].end,$
            \Statex \qquad \qquad \qquad $scene\_start,$
            \Statex \qquad \qquad \qquad $\textit{speech\_segments}[idx].start - \textit{segment\_duration}/2 $
            \Statex \qquad \qquad \qquad $\big)$
            \State $window\_start \gets \text{RandomUniform}(lower\_bound, upper\_bound)$
        \EndIf

        \State $window\_end \gets window\_start + \textit{segment\_duration}$
        \If{$window\_end \ge scene\_end$}
            \State \textbf{break}
        \EndIf

        \State append $(window\_start, window\_end)$ to $\textit{speech\_single\_shot\_segments}$
        \State $idx \gets$ next $idx$ such that $\textit{speech\_segments}[idx].start > window\_end$
    \EndWhile

    \State $i \gets i + 1$
\EndWhile

\end{algorithmic}
\end{algorithm}

\subsection{Detailed Filtering Thresholds}
\label{app:thresholds}

In this section, we provide the specific filtering thresholds used during our data quality assessment process, as discussed in Sec.~\ref{sec:quality_assess}. These thresholds were determined by empirical observation to ensure a high-quality corpus while maintaining sufficient data diversity.

\subsubsection{Threshold Specifications}

Table~\ref{tab:thresholds} summarizes the metrics and their corresponding cutoffs across the three dimensions: audio quality, video quality, and audio-visual alignment.

\begin{table}[ht]
\centering
\caption{Filtering thresholds for data quality assessment.}
\label{tab:thresholds}
\begin{tabular}{@{}llc@{}}
\toprule
\textbf{Dimension} & \textbf{Metric} & \textbf{Threshold} \\ \midrule
\multirow{5}{*}{Audio Quality} & Silence Ratio & $< 0.8$ \\
 & Bandwidth (Hz) & $> 1,000$ \\
 & Audiobox-PQ (Production Quality) & $> 5.0$ \\
 & Audiobox-CU (Content Usefulness) & $> 4.5$ \\
 & Audiobox-CE (Content Enjoyment) & $> 2.5$ \\ \midrule
\multirow{2}{*}{Video Quality} & DOVER-Aesthetic & $> 0.85$ \\
 & DOVER-Technical & $> 0.05$ \\ \midrule
\multirow{2}{*}{A-V Alignment} & ImageBind Score (IB-Score) & $\geq 0.2$ \\
 & \textbf{OR} SynchFormer Offset (DeSync) & $\leq 0.5$ \\ \bottomrule
\end{tabular}
\end{table}

\subsubsection{Subset Construction and Logic}

\begin{itemize}
    \item \textbf{Audio-Visual Alignment Logic}: For alignment filtering, we apply a relaxed logical ``OR'' gate between semantic and temporal alignment. A video is retained if it satisfies either $\text{IB-Score} \geq 0.2$ \textbf{OR} $\text{DeSync} \leq 0.5$. This ensures that both semantically relevant ambient sounds and temporally synchronized speech/actions are preserved.
    
    \item \textbf{Speech Data Filtering}: To construct specialized subsets for tasks such as lip synchronization, we utilize the EAT \cite{chen2024eat} classification model. Specifically, for the \textit{Speech Subset}, we only retain samples where both \texttt{EAT-contained-Speech} and \texttt{EAT-contained-Singing} tags are evaluated as \texttt{True} (or satisfy the model's positive classification confidence).
\end{itemize}

\subsection{Audio-visual Captioning Details}
\label{app:caption}
We present detailed prompt design schemes tailored for our multimodal annotation pipeline, comprising three key components: (1) prompts for video captioning utilizing the MiMo-VL-7B-RL model \cite{Yue2025MiMoVLTR}; (2) prompts for the dual-model audio strategy, specifically covering speech transcription with Qwen3-Omni-Instruct \cite{Qwen3-Omni} and non-speech sound captioning with Qwen3-Omni-Captioner \cite{Qwen3-Omni}; and (3) instructions for the GPT-OSS-120B model \cite{agarwal2025gpt} to facilitate consistency checks and caption merging.

\begin{promptblock}
\setlength{\parskip}{0.2cm}
\linespread{\normalbaselineskip \baselineskip}
\textbf{Visual Description Prompt}

You are a \textbf{visual description annotator}. Your sole purpose is to produce event-level annotations \textbf{only for verifiably visible content} (objects, actions, scenes, text). Ignore audio and inferred context entirely!

\textbf{LAWS}

\begin{enumerate}
    \item \textbf{LAW OF VISUAL TRUTH}
    \begin{itemize}
        \item Only describe \textbf{visually verifiable elements} (objects, actions, scenes, text).
        \item Do \textbf{not} infer or hallucinate based on audio or context.
    \end{itemize}

    \item \textbf{LAW OF VISUAL SILENCE}
    \begin{itemize}
        \item When no visual change is present, output \texttt{null} for visual\_description.
        \item When no on-screen text is present, output \texttt{null} for on\_screen\_text.
    \end{itemize}

    \item \textbf{LAW OF VISUAL DYNAMICS}
    \begin{itemize}
        \item Detect all transitions: emerging/ceasing/moving visual elements.
        \item Precisely document motion trajectories, speed changes, and visual rhythm.
    \end{itemize}
\end{enumerate}

\noindent\rule{\linewidth}{0.4pt}

\textbf{OUTPUT FORMAT}

\noindent\texttt{%
\{\\
\hspace*{1em}"video\_visual\_report": \{\\
\hspace*{2em}"visual\_description": "[Describe how visual elements evolve in detail.\\
\hspace*{2em}Ignore all text, subtitle and watermark in the video.\\
\hspace*{2em}\texttt{null} if no visual change is present]",\\
\hspace*{2em}"on\_screen\_text": "[Exact transcription of all visible text.\\
\hspace*{2em}\texttt{null} if no text is visually present]"\\
\hspace*{1em}\},\\
\hspace*{1em}"final\_verification\_audit": \{\\
\hspace*{2em}"hallucination\_check\_passed": true,\\
\hspace*{2em}"visual\_changes\_verified": true,\\
\hspace*{2em}"comment": "All visual motion dynamics verifiably detected.\\
\hspace*{2em}No audio content or contextual inference included."\\
\hspace*{1em}\}\\
\}%
}

\end{promptblock}

\begin{promptblock}
\setlength{\parskip}{0.2cm}
\linespread{\normalbaselineskip \baselineskip}
\textbf{Audio Captioner Prompt}

Please describe the audio you hear in detail.

\end{promptblock}

\begin{promptblock}
\setlength{\parskip}{0.2cm}
\linespread{\normalbaselineskip \baselineskip}
\textbf{Speech Transcription Prompt}

You are a \textbf{speech transcription annotator}. Your task is Automatic Speech Recognition! Your sole purpose is to produce event-level \textbf{verbatim transcriptions of audible speech}. Ignore non-speech sounds and music entirely.
You must follow the OUTPUT FORMAT!

\textbf{LAWS}

\begin{enumerate}
    \item \textbf{LAW OF LANGUAGE FIDELITY}
    \begin{itemize}
        \item Preserve the original language.
        \item No translation.
    \end{itemize}

    \item \textbf{LAW OF SPEECH DYNAMICS}
    \begin{itemize}
        \item When new speaker/language/intonation begins, create a new event. But you still need to follow the output format.
    \end{itemize}

    \item \textbf{LAW OF SILENCE}
    \begin{itemize}
        \item When no speech is present, output \texttt{null} in `speech\_description` part.
    \end{itemize}
\end{enumerate}

\noindent\rule{\linewidth}{0.4pt}

\textbf{OUTPUT FORMAT}

\noindent\texttt{%
\{\\
\hspace*{1em}"speech\_transcription\_report": \{\\
\hspace*{2em}"speech\_description": "[Exact verbatim speech. \texttt{null} if none.\\
\hspace*{2em}Use [inaudible] for unclear parts]"\\
\hspace*{1em}\},\\
\hspace*{1em}"final\_verification\_audit": \{\\
\hspace*{2em}"hallucination\_check\_passed": true,\\
\hspace*{2em}"speech\_dynamics\_verified": true,\\
\hspace*{2em}"comment": "All speech changes verifiably detected.\\
\hspace*{2em}No non-speech or music included."\\
\hspace*{1em}\}\\
\}%
}

\end{promptblock}

\begin{promptblock}
\setlength{\parskip}{0.2cm}
\linespread{\normalbaselineskip \baselineskip}
\textbf{Merge Caption Prompt}

\textbf{Act as a video description consolidator.} Your task is to merge the provided \texttt{video\_description}, \texttt{audio\_description}, and \texttt{speech\_description} fields into a single, fluent, and natural-sounding paragraph that vividly conveys the full multimedia experience to someone who cannot see or hear the original. Follow these rules precisely:

\begin{enumerate}
    \item \textbf{Start with the visual narrative and anchor speaker context}
    \begin{itemize}
        \item If \texttt{video\_description} is provided, begin with it, weaving in visual cues that ground speaker identities or positions (e.g., "A woman in a red coat stands by the door; nearby, a man leans against the counter").
        \item Describe all visual content in specific, chronological, natural language—actions, objects, settings, lighting, movements, and scene changes—without summarizing. Use these details to contextualize speakers (e.g., "The child pointing at the painting" or "The older man gesturing emphatically").
    \end{itemize}

    \item \textbf{Integrate speech as visually anchored dialogue with speaker dynamics}
    \begin{itemize}
        \item If \texttt{speech\_description} is provided, embed spoken words as \textbf{quoted dialogue} tied to visual elements, using speech verbs that reflect tone (e.g., "snaps," "murmurs," "laughs") and speech rate (e.g., "rushes," "drawls," "pauses between words").
        \item Use \texttt{audio\_description} to inform:
        \begin{itemize}
            \item Total number of speakers (e.g., "two distinct voices" or "a group of overlapping speakers").
            \item Speaker transitions, paired with visual cues when possible (e.g., "As the camera pans to the window, a new voice cuts in sharply: 'Wait, that's not right'").
            \item Voice characteristics (e.g., "high-pitched," "gruff") to align with visible subjects (e.g., "the teenager in the corner" or "the gray-haired woman").
        \end{itemize}
        \item Example: \textit{"The man slams his fist on the table. 'I told you this would happen!' he shouts. Across the room, the woman in glasses crosses her arms and replies slowly, 'You never listened to my warnings.'"}
        \item Rely strictly on \texttt{speech\_description} for the actual spoken content; use \texttt{audio\_description} for speaker dynamics and \texttt{video\_description} to anchor them visually.
    \end{itemize}

    \item \textbf{Describe non-speech audio concisely and distinctly}
    \begin{itemize}
        \item If \texttt{audio\_description} is provided, include \textbf{only non-speech elements}, introduced with "The audio includes…" or "In the background…".
        \item Limit this to four categories:
        \begin{itemize}
            \item Ambient/environmental sounds (e.g., rain tapping the window, a crowd's murmur);
            \item Music (e.g., a soft piano melody, upbeat jazz);
            \item Audio themes/sources (e.g., a fire alarm blaring, waves crashing);
            \item Structural audio changes (e.g., silence giving way to a crescendo).
        \end{itemize}
        \item Omit all references to voices, words, or speech—these belong exclusively to the speech section.
    \end{itemize}

    \item \textbf{Ensure coherence and avoid repetition}
    \begin{itemize}
        \item Never duplicate details across sections (e.g., don't restate a speaker's visual position in the audio section).
        \item Link elements logically: e.g., a visual action (a character gasping) paired with speech ("'Oh no!' she exclaims") and ambient sound (a glass shattering in the background).
    \end{itemize}

    \item \textbf{Omit empty fields and maintain fluency}
    \begin{itemize}
        \item Skip missing/empty fields.
        \item The final output should read as a seamless, human-like narrative—prioritizing flow, sensory immersion, and clarity over rigid structure.
    \end{itemize}
\end{enumerate}

\end{promptblock}

\begin{figure}[htbp]
  \centering
  % 为了好调，定义单张小图的宽度（6 张 + 5 个 \hfill ≈ 占满一行）
  \newcommand{\imgw}{0.166\linewidth}

  % —— 第一排：6 张 —— 
  \begin{subfigure}[t]{\imgw}
    \centering
    \includegraphics[width=\linewidth]{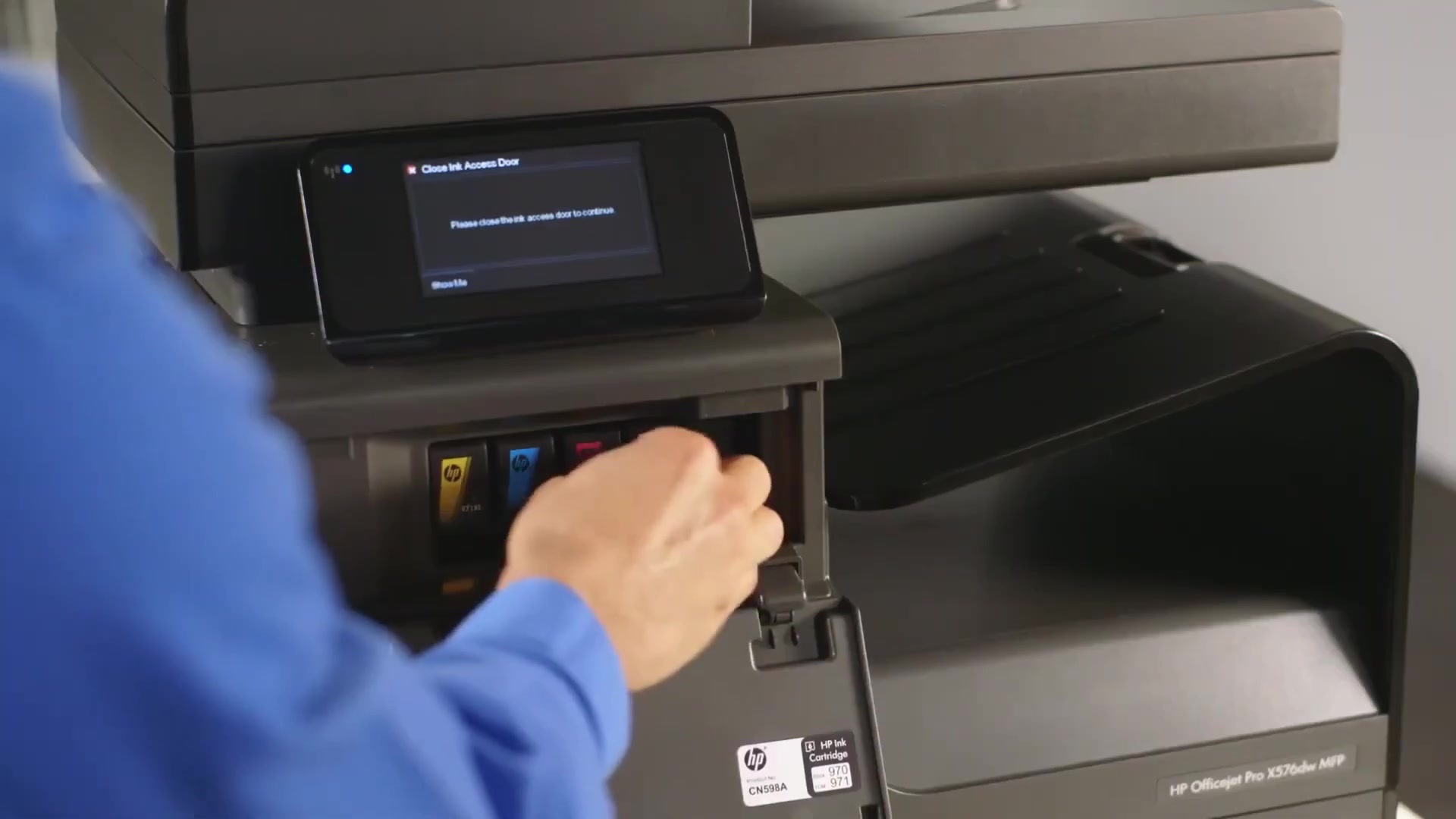}
  \end{subfigure}\hfill
  \begin{subfigure}[t]{\imgw}
    \centering
    \includegraphics[width=\linewidth]{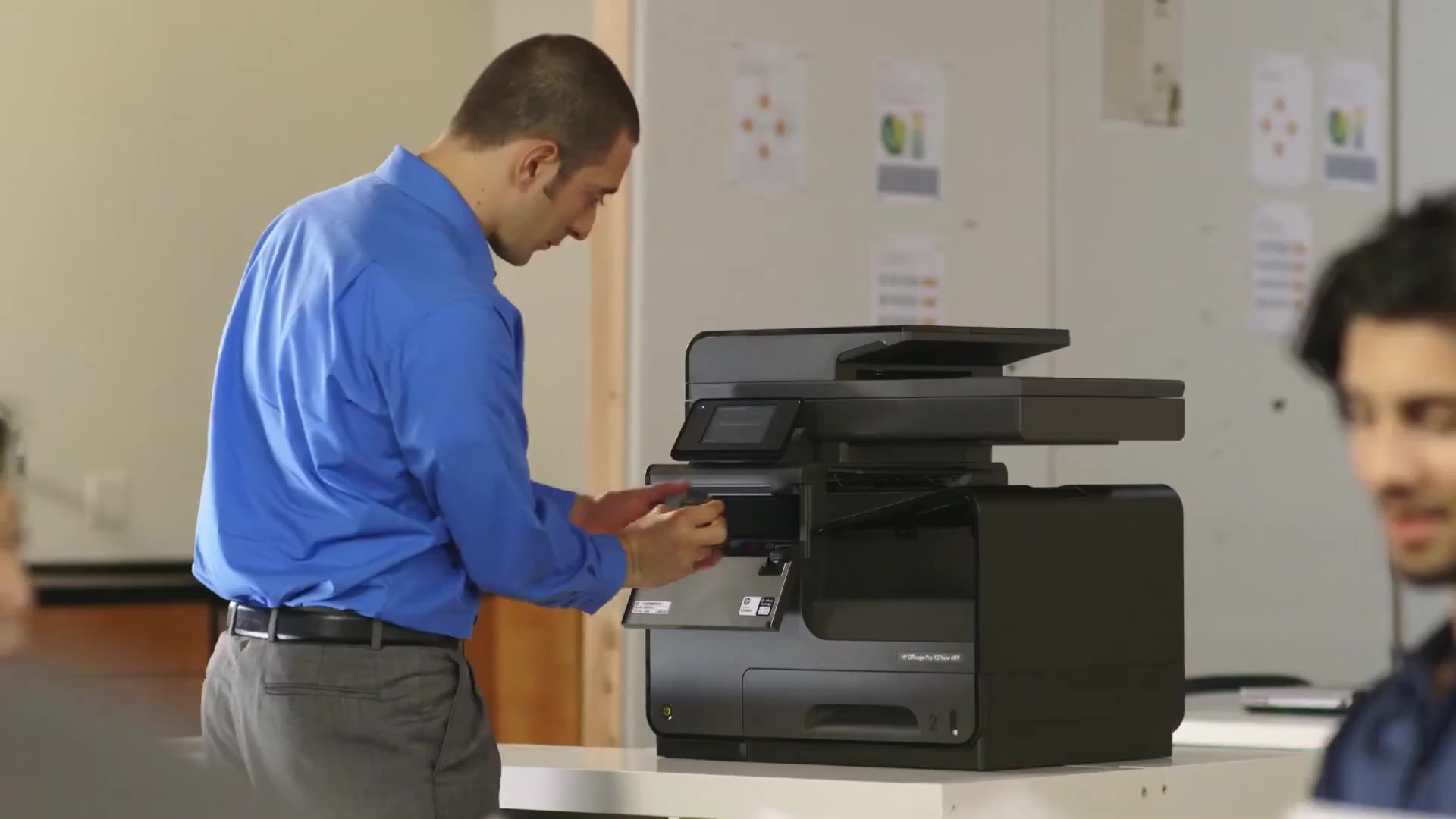}
  \end{subfigure}\hfill
  \begin{subfigure}[t]{\imgw}
    \centering
    \includegraphics[width=\linewidth]{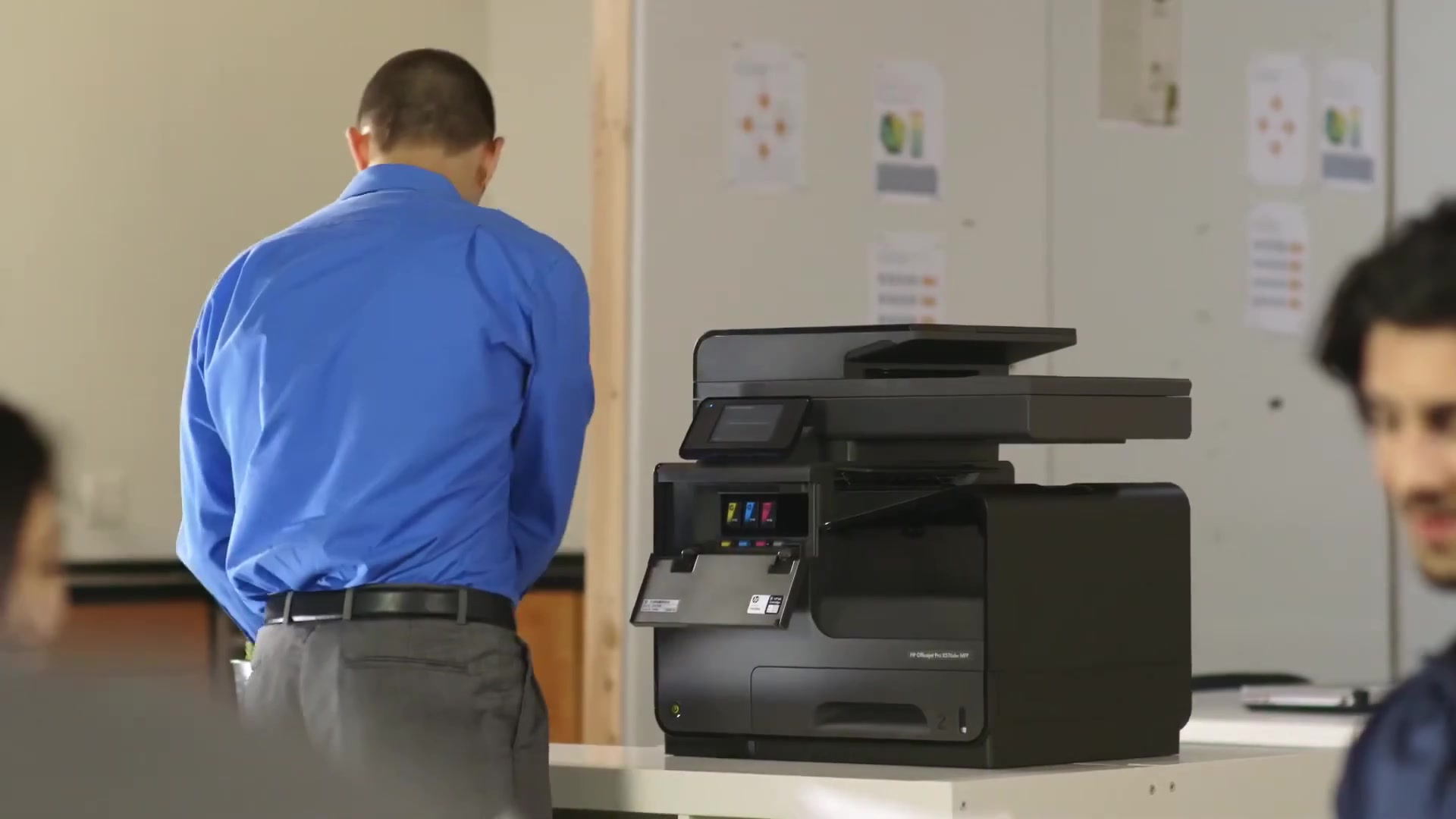}
  \end{subfigure}\hfill
  \begin{subfigure}[t]{\imgw}
    \centering
    \includegraphics[width=\linewidth]{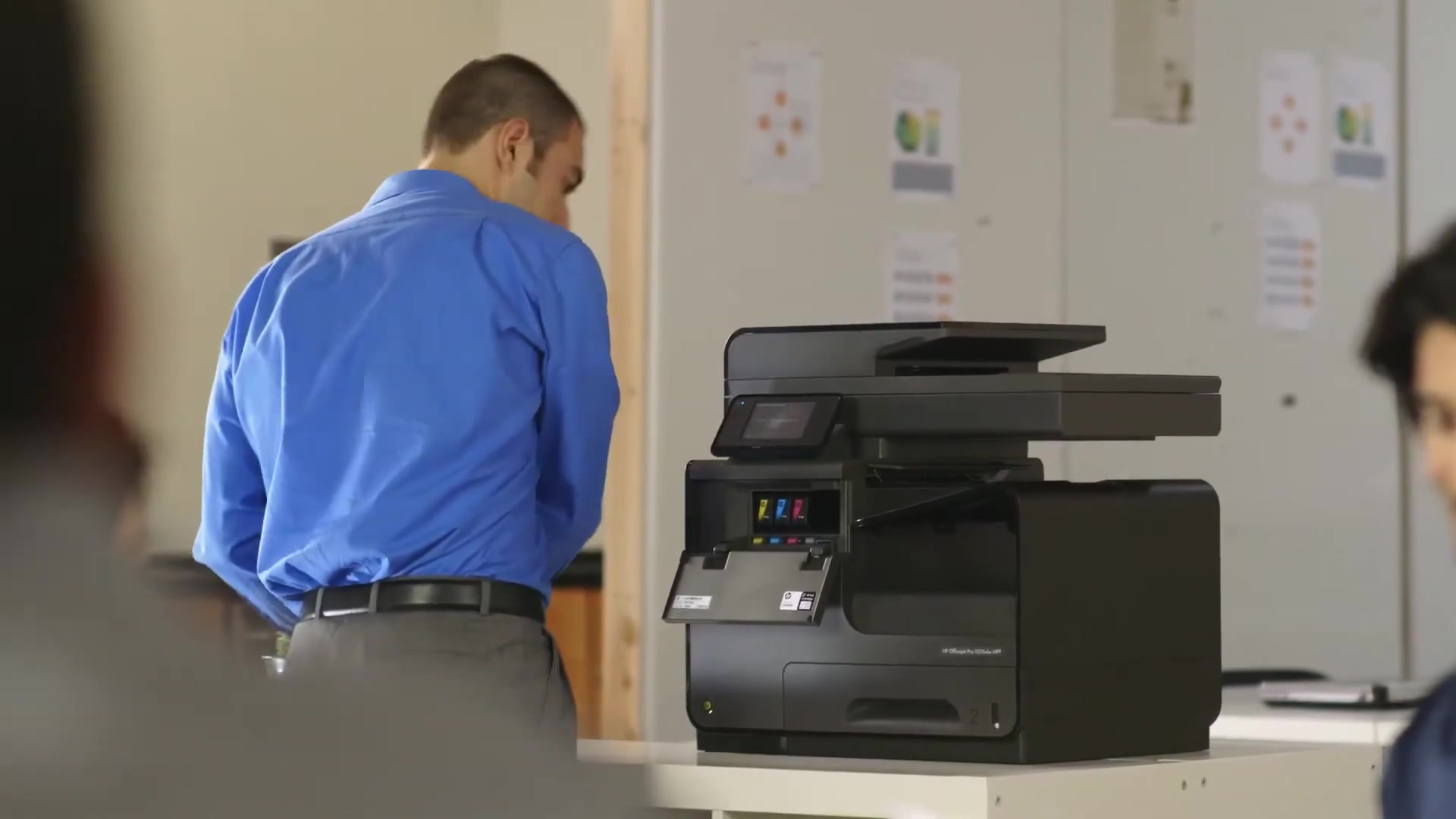}
  \end{subfigure}\hfill
  \begin{subfigure}[t]{\imgw}
    \centering
    \includegraphics[width=\linewidth]{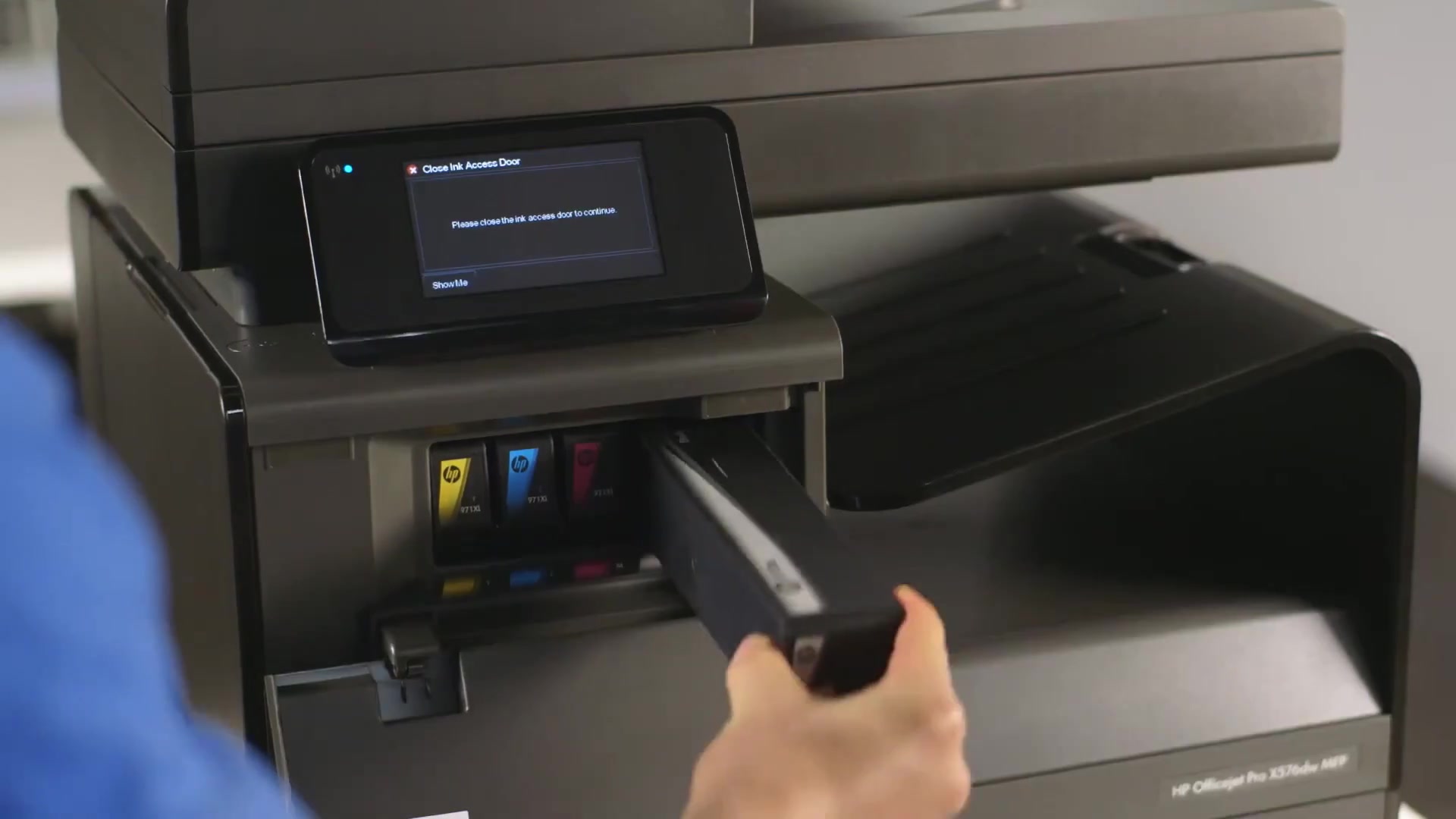}
  \end{subfigure}\hfill
  \begin{subfigure}[t]{\imgw}
    \centering
    \includegraphics[width=\linewidth]{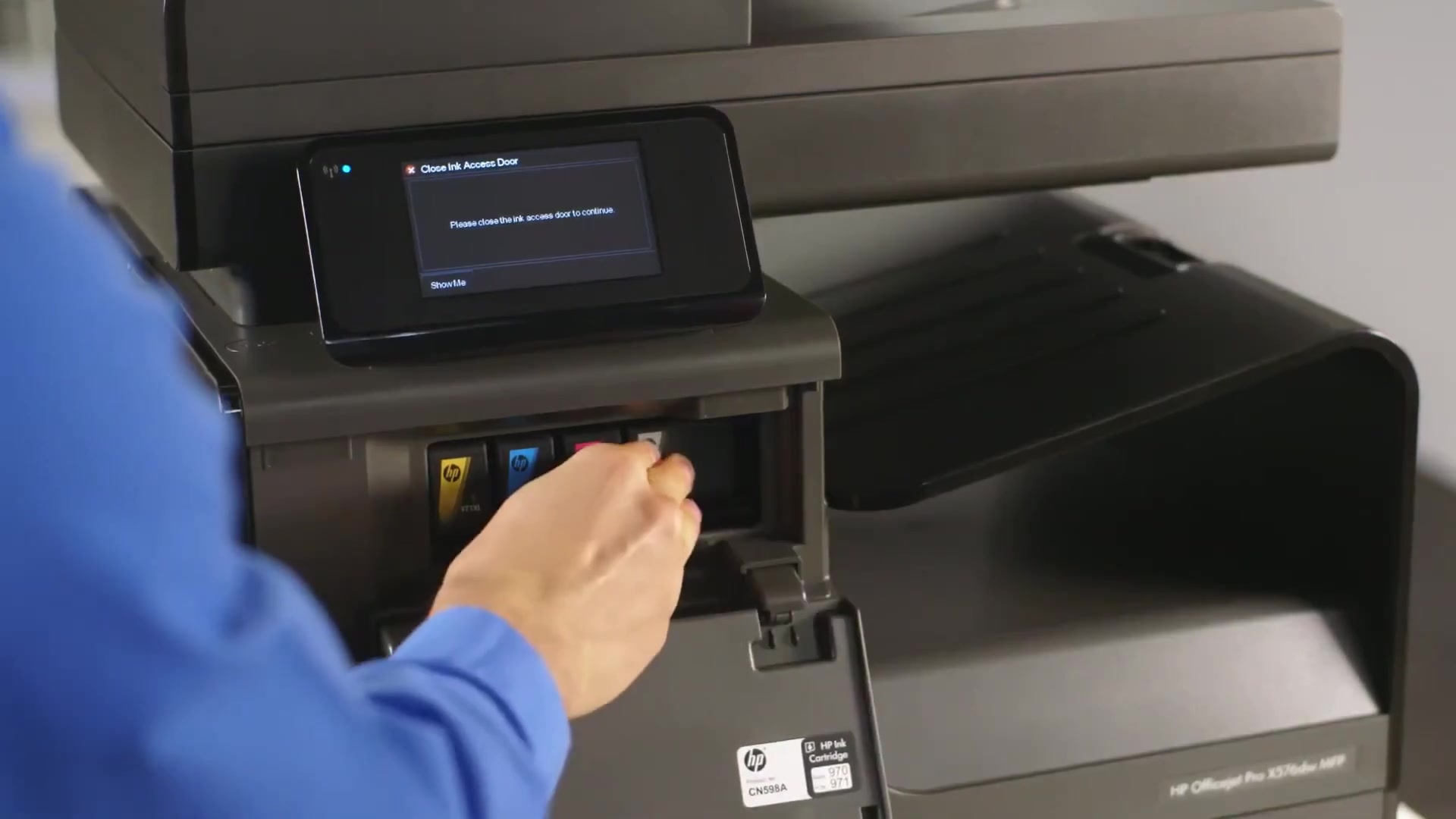}
  \end{subfigure}

  \vspace{0.05em} % 行间距，可按需调整

  % —— 第二排：1 张，铺满整行 —— 
  \begin{subfigure}[t]{\linewidth}
    \centering
    \includegraphics[width=\linewidth]{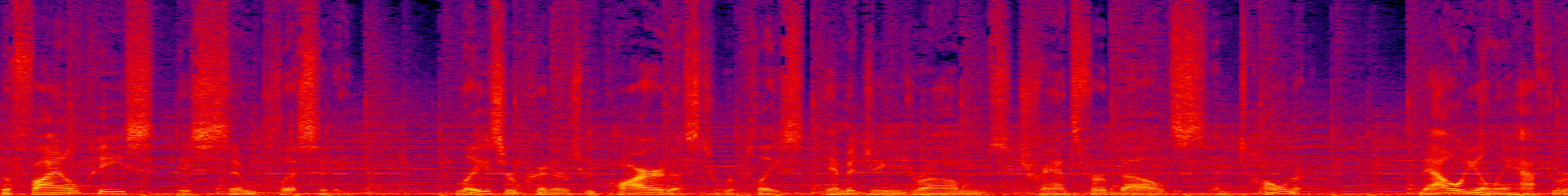}
  \end{subfigure}

  \caption{Example video 1 for audio-visual captioning.}
  \label{fig:caption_eg1}
\end{figure}

We provide the caption examples generated by bimodality models, including the caption results of three dimensions: sound, music, and video, as well as the final total caption result, which is obtained by merging the dimension-specific captions mentioned above using a large language model \cite{agarwal2025gpt}. \Cref{fig:caption_eg1} presents the original video clips and audio mel-spectrogram corresponding to this example.

\begin{promptblock}
\setlength{\parskip}{0.2cm}

\textbf{Visual Description}

A close-up of an HP OfficeJet Pro X576dw MFP printer shows its ink cartridge bay with yellow (Y 971XL), blue (C 971XL), magenta (M 971XL), and an open black (X) slot. A person in a blue shirt and gray pants holds a black HP ink cartridge (labeled 970/971) and moves it toward the black slot. The printer's display reads, ``Close Ink Access Door'' and ``Please close the ink access door to continue.'' A wider shot reveals the person from the back/side, continuing to align and insert the black cartridge into the X slot, with an office background (posters, a table, and blurred figures) visible. The camera alternates between close-ups of the hand/printer interaction and wider office-scene shots, documenting the cartridge insertion motion. After insertion, the hand finishes positioning the cartridge into the slot.

\textbf{Audio Description}

The audio clip is a tightly produced, 8-second segment designed for commercial or corporate advertising, likely for a laser printer product or service. The recording quality is high, with a full frequency spectrum, no distortion, and a polished, modern sound profile. A single, mature male voice---clear, authoritative, and with a General American accent---delivers the narration in a calm, professional tone. The content is as follows:

\begin{itemize}
    \item ``The final piece is simplicity.''
    \item ``Laser printers required us to replace imaging drums, transfer belts, and toner.''
    \item ``Now we only have to---''
\end{itemize}

The narration is delivered in a measured, declarative style, with a subtle rise in pitch at ``simplicity'' to signal a transition to a new point. The phrase ``the final piece'' is spoken with a slight emphasis, and the listing of printer components (``imaging drums, transfer belts, and toner'') is given with a rhythmic, evenly paced cadence. The concluding phrase ``Now we only have to---'' is abruptly cut off, suggesting the audio is an excerpt from a longer ad or commercial.

Throughout, a subtle, unobtrusive electronic background music track plays. It features synthesized pads, a soft, steady beat, and a simple melodic line, all in a minor key. The music is mixed low and never interferes with the narration, serving to create a modern, professional, and slightly optimistic atmosphere.

The recording is studio-clean, with no background noise or environmental sounds, and the voice has a slight reverb, indicating a controlled studio environment. The overall production style, linguistic choices, and content are all consistent with modern North American advertising aimed at businesses or professional consumers, highlighting the technological advancement and ease-of-use of a new printer or printer service. The abrupt ending reinforces that this clip is a fragment of a larger narrative, likely intended to leave a strong impression of progress and simplicity before the listener's attention is drawn elsewhere.

\textbf{Speech Description}

The final piece is simplicity. Laser printers required us to replace imaging drums, transfer belts, and toner. Now we only have to...

\noindent\rule{\linewidth}{0.4pt}

\textbf{Integrated Caption}

A close-up reveals the ink cartridge bay of an HP OfficeJet Pro X576dw MFP, its yellow, blue and magenta cartridges neatly seated while the black slot hangs open; the printer's display flashes the messages ``Close Ink Access Door'' and ``Please close the ink access door to continue.'' A person wearing a blue shirt and gray pants reaches in, holding a black HP ink cartridge labeled 970/971, and carefully aligns it with the empty X slot. The camera alternates between tight shots of the hand guiding the cartridge and wider views of the office, where posters, a table and blurred figures form the background, before the hand slides the cartridge fully into place. As this insertion unfolds, a calm, authoritative male voice---off-screen, mature and speaking with a clear General American accent---states, ``The final piece is simplicity.'' He continues, ``Laser printers required us to replace imaging drums, transfer belts, and toner,'' delivering the list with an even, rhythmic cadence, then adds, ``Now we only have to---'' before the narration is abruptly cut off. The audio includes a subtle, unobtrusive electronic background music track of synthesized pads, a soft steady beat and a simple melodic line in a minor key, mixed low so it underlies the narration without interfering.
\end{promptblock}

\subsection{Benchmark Details}
\label{app:benchmark_details}

This appendix provides additional details of the benchmark construction. Our benchmark is constructed from real-world videos by extracting the first-frame image and a corresponding prompt. Each prompt concisely describes key visual elements, such as the scene setting, characters, and environmental conditions. Depending on the scenario, audio-related information is incorporated to form a unified prompt for joint video--audio generation. All samples are adapted to ensure temporal consistency and logical coherence of the generated videos.

The benchmark consists of six scenario categories, each targeting specific challenges in joint video--audio generation:
\begin{itemize}
    \item \textbf{Multi-speaker scenarios}, which evaluate the model’s ability to generate synchronized speech, facial expressions, and interactions among multiple characters.
    \item \textbf{Movie videos}, which require film-level narrative generation with plots referencing the background of the original films.
    \item \textbf{Sports competitions}, focusing on athletes’ performances, with some prompts including commentators’ narration.
    \item \textbf{Game livestream videos}, covering shooting games, 3D games, and competitive games.
    \item \textbf{Camera motion sequences}, designed to assess visual realism under camera panning, zooming, and rotation.
    \item \textbf{Anime-style videos}, including both 2D anime and 3D animated content.
\end{itemize}

\end{document}